\documentclass[12pt, a4paper]{article}

\usepackage{float}
\usepackage{soul}
\usepackage{comment}
\usepackage{pgfplots}
\usepackage{tikz}
\usetikzlibrary{positioning, arrows, shapes}
\usepackage{float}
\usepackage{amsmath}
\usepackage{amssymb}
\usepackage{graphicx}
\usepackage{tabularx}
\usepackage{mathrsfs}
\usepackage{longtable}
\usepackage{multirow}
\usepackage{amsmath}
\usepackage{amsfonts}
\usepackage{bbm}
\usepackage{times}
\usepackage{etoolbox}
\usepackage{booktabs}
\usepackage{enumitem}
\usepackage{url}
\usepackage[round]{natbib}
\usepackage[boxed,algoruled]{algorithm2e}
\usepackage{algpseudocode}

\newtheorem{example}{Example}

\def\mN{\mathcal{N}}

\def\mM{\mathcal{M}}
\def\mA{\mathcal{A}}
\def\mG{\mathcal{G}}
\def\fringe{\mathit{fringe}}

\newcommand{\AlgorithmComment}[1]{{\scriptsize{\tcp*[h]{#1}}}}
\newcommand{\IfLine}[2]{\textbf{if} #1 \textbf{then} #2}
\newcommand{\ElseLine}[1]{\textbf{else} #1}
\newcommand{\ForLine}[2]{\textbf{for} #1 \textbf{do} #2}

\newcommand{\dist}{\mathit{dist}}

\newcommand{\CS}{\mathit{CS}}
\newcommand{\MC}{\mathit{MC}}

\newcommand{\set}[1]{\{#1\}}  
\newcommand{\midd}{\mathrel{:}}

  \newlength{\wordlength}

\newcommand{\citefull}[1]{\citeauthor{#1} \cite{#1}}

\newtheorem{definition}{Definition}

\begin{document}



\title{Game-theoretic Network Centrality: A Review}

\author{Mateusz K. Tarkowski,\\Tomasz P. Michalak, Talal Rahwan, Michael Wooldridge}
\maketitle

\begin{abstract}
Game-theoretic centrality is a flexible and sophisticated approach to identify the most important nodes in a network. It builds upon the methods from cooperative game theory and network theory. The key idea is to treat nodes as players in a cooperative game, where the value of each coalition is determined by certain graph-theoretic properties. Using solution concepts from cooperative game theory, it is then possible to measure how responsible each node is for the worth of the network. 

The literature on the topic is already quite large, and is scattered among game-theoretic and computer science venues. We review the main game-theoretic network centrality measures from both bodies of literature and organize them into two categories: those that are more focused on the connectivity of nodes, and those that are more focused on the synergies achieved by nodes in groups. We present and explain each centrality, with a focus on algorithms and complexity.
\end{abstract}

\section{Introduction}

\noindent In many social networks, certain nodes play more important roles than others. For example, popular individuals with frequent social contacts are more likely to spread a disease in the event of an epidemic \cite{Dezso:Barabasi:2002}; airport hubs such as Heathrow or Schiphol concentrate intercontinental passenger traffic \cite{Adler:2001}; and certain parts of the brain's neural network may be indispensable for breathing or to perform other fundamental actions \cite{Keinan:2004}. As a result, the concept of \emph{network centrality}, which aims to quantify the importance of nodes and edges, has been extensively studied in the literature \cite{Koschutzki:et:al:2005,Brandes:Erlebach:2005}. 

\begin{figure}[thbp]
\center
\includegraphics[width=7.5cm]{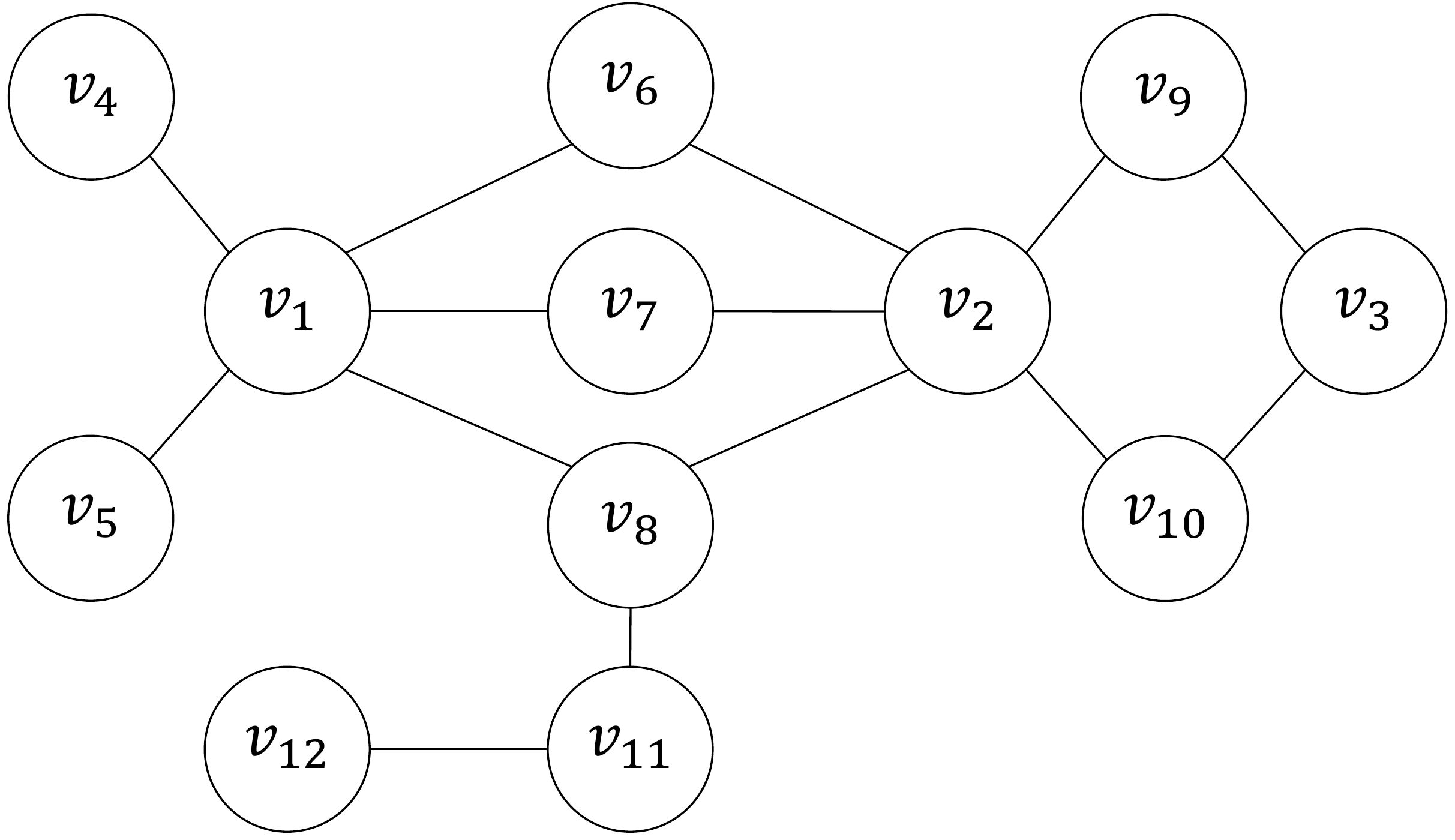}
\caption{Sample network of 12 nodes. The top nodes are $v_1$ and $v_2$ according to degree centrality, $v_8$ according to closeness centrality, $v_2$ according to betweenness centrality, and again $v_2$ according to eigenvector centrality.
}
\label{fig:example}
\end{figure}

A number of centrality measures have been proposed in the literature, and among the best known of these are degree, closeness, betweenness, and eigenvector centralities \cite{Freeman:1979,Bonacich:1972}.
Degree centrality quantifies the power of a node by the number of its incident edges. For example, nodes $v_1$ and $v_2$ in the network in Figure~\ref{fig:example} have degree 5 and, from the perspective of degree centrality, these are the key nodes in the network. On the other hand, closeness centrality promotes nodes that are close to all other nodes in the network (i.e., nodes from which it is possible to reach other nodes in a smaller number of steps are ranked higher). According to this measure, node $v_8$ in Figure~\ref{fig:example} is ranked highest. Next, betweenness centrality counts shortest paths (i.e., paths that use the minimal number of links) between any two nodes in the network and ranks nodes according to the number of shortest paths they belong to. With this ranking, $v_2$ in Figure~\ref{fig:example} becomes the top node. Finally, eigenvector centrality is based on the idea that connections to more important nodes should be valued more than otherwise equal connections to less important nodes~\cite{Bonacich:1972}. In this case, node $v_2$ in Figure~\ref{fig:example} is again ranked highest.

Since real-world networks have certain specific features, and there are various perspectives from which they need to be analyzed, several extensions to the above classical measures have been proposed in the literature. One such extension is the concept of \textit{group centrality} introduced by~\cite{Everett:Borgatti:1999}, where the notions of classical centrality measures are extended to groups of nodes instead of only individual nodes. By doing so, it is possible to capture \textit{synergies} that emerge when the roles of nodes are considered jointly. 
To illustrate this, let us assume that the network from Figure~\ref{fig:example} represents districts in a city, and edges are communication routes. Our aim is to identify locations for two hospitals, so that they can be reached as quickly as possible from any district in the city. Taking the two top nodes from the ranking obtained using closeness centrality (i.e., $v_8$ and $v_2$) in this case does not lead to an optimal outcome, since those two nodes are adjacent to each other. On the other hand, considering the joint closeness centrality of all pairs of nodes is the better approach, since $\{v_1, v_2\}, \{v_1, v_3\}, \{v_1, v_9\}, $ or $\{v_1, v_{10}\}$ each constitute an optimal outcome.

Although the concept of group centrality addresses the issue of synergy between the nodes when their role is considered jointly, it suffers from a fundamental deficiency. There is an exponential number of such groups and, even if one could examine all of them, it is not clear how to construct a consistent ranking of \emph{individual} nodes using the group results. Specifically, should the nodes from the most valued groups be ranked highest? Or should we focus on nodes from the group with the highest average performance per node? Or maybe the nodes which contribute most to every group they join are the best? In fact, there are very many possibilities to choose from.

In this paper we discuss the existing literature on game-theoretic network centrality---a recent research direction that provides a compelling answer to the above questions by applying techniques rooted in cooperative game theory.\footnote{\footnotesize We note that there exists a literature on a game-theoretic approach to network centrality based on \emph{non-cooperative} game theory, e.g., \cite{Calvo:Armengol:2001}. Nevertheless, it is customary in the literature to use the term \textit{game-theoretic network centrality} for techniques pertaining to \emph{cooperative} game-theory applied to the analysis of node centrality in networks. We will follow this convention also in this article.} The literature on the topic has been expanding over the past years, and is scattered among game-theoretic and computer science venues.
We identify two separate lines of research in game-theoretic network centrality. The first 
is primarily concerned with applying existing and developing new cooperative solution concepts that essentially depend on an underlying network structure. In other words, these solution concepts can be applied only to networks and are not well-defined otherwise. We refer to the centrality measures from this line of research as game-theoretic measures based on connectivity, since the solution concepts that it is concerned with typically attempt to account for the role played by nodes in connecting otherwise disconnected groups in the network. The Myerson value \cite{Myerson:1977} is probably the most well-known such concept.

On the other hand, the second line of research is concerned with applying (both new and existing) cooperative solution concepts for regular cooperative games to specific classes of cooperative games defined on networks through group centrality measures. 
We refer to the measures from this line of research as game-theoretic centrality measures not based on connectivity.

In what follows, we present a comprehensive review of both types of game-theoretic centrality, with particular emphasis on computational complexity. Section~\ref{ch:3-preliminaries} introduces necessary notation and definitions. In Section~\ref{section:definition}, we define the concept of game-theoretic network centrality and show the difference between this approach to centrality and the classic ones. Section~\ref{section:connectivity} deals with game-theoretic network centrality measures based on connectivity, while Section~\ref{section:synergy} deals with those that are not. In Section~\ref{section:applications} we present some applications of game-theoretic network centrality considered in the literature. Conclusions follow.

\section{Preliminaries}\label{ch:3-preliminaries}

In this section, we will introduce the key concepts pertaining to cooperative game theory and graph theory.

\subsection{Game-Theoretic Concepts}\label{section:gameTheoreticConcepts}
\noindent We first discuss the basic model of cooperative games, its generalisation, the concepts of the unanimity basis and dividends. Next, we introduce a number of solution concepts in cooperative game theory---the cornerstone on which all the game-theoretic network centralities discussed in this article are built.

\paragraph{Cooperative games:} A cooperative game consists of a set of \textit{players}, $I = \{1, 2, \ldots , n\}$, and a \emph{characteristic function}, $\nu: 2^I \rightarrow \mathbb{R}$, that assigns to each \emph{coalition} $C \subseteq I$ of players a real value indicating its performance (we assume $\nu(\emptyset) = 0$). A cooperative game in characteristic function form, then, is a pair $(I, \nu)$. Following convention, we will refer to such a game by $\nu$, omitting the set of players.




\paragraph{Types of Cooperative Games: } Let us single out the following types of cooperative games:

\begin{itemize}
	\item $\nu$ is \textit{superadditive} if and only if $\nu(S \cup T) \geq \nu(S) + \nu(T)$ for all $S,T\subseteq I$ such that $S\cap T=\emptyset$. This means that when two disjoint coalitions join, their worth is at least the sum of their values. In other words, there is no negative synergy;
	
	\item $\nu$ is \textit{convex} if and only if $\nu(S \cup T) + \nu(S \cap T) \geq \nu(S) + \nu(T)$ for all $S,T\subseteq I$. This property captures the ``snowballing'' effect. In other words, the incentive to join a coalition rises as the coalition increases in size; and
	
	\item $\nu$ is \textit{symmetric} if and only if there exists a function $f$ such that $\nu(S) = f(|S|)$ for all $S \subseteq I$. That is, the value of a coalition depends only on its size, and not on the individual properties of its members.
\end{itemize}
Convex games were defined by \cite{Shapley:1971}. Superadditive and symmetric games are also discussed in this paper.

\paragraph{Unanimity Basis and Harsanyi Dividends for Cooperative Games:} It is possible to express any characteristic function  $\nu$ as a linear combination of the \textit{unanimity basis} for $I$. For every set of players $I$ there exists a unanimity basis consisting of $2^{n} - 1$ \textit{unanimity games} (one for each subset of $I$, excluding the empty set $\emptyset$), where $n$ is the number of players in $I$. A unanimity game, $u_S$ (where $S \subseteq I$), has the following form:
\[
u_S(C) = \begin{cases} {1} {\text{   if } S \subseteq C} \\ {0} {\text{   otherwise}} \end{cases}.
\]
In other words, we say a coalition $C$ is \emph{winning} if it contains all of the players in $S$, and \emph{losing} otherwise. It is not difficult to see that the unanimity games do indeed form a basis. Since the space of characteristic functions with $n$ players can be interpreted as a ($2^n - 1$)-dimensional vector space, and it is easy to check that each of the $2^n - 1$ elements in the unanimity basis is linearly independent, then it follows that the unanimity basis has to span the whole space.

For any cooperative game, we call its coordinates in the unanimity basis the \textit{Harsanyi Dividends} \cite{harsanyi:1958}, where the coordinate of $u_s$ is denoted by $\Delta_{\nu}(S)$. This results in the following:
\begin{equation}
\nu(C) = \sum_{S \subseteq I} \Delta_{\nu}(S) u_S(C).
\end{equation}
%
It can be proven that:
\begin{equation}
\Delta_{\nu}(S) = \sum_{T \subseteq S} (-1)^{|S| - |T|} \nu(T).
\end{equation}
Perhaps a more illuminating (albeit recursive) definition of Harsanyi dividends is as follows:
\begin{equation}\label{equation:harsanyi:recursive}
\Delta_{\nu}(S) = \nu(S) - \sum_{T \subset S} \Delta_{\nu}(T).
\end{equation}

\paragraph{Solution Concepts for Cooperative Games:} Assuming that the \emph{grand coalition}---the coalition consisting of all players in the game---is formed, and assuming that it has the highest value\footnote{\footnotesize Of course, there exist solution concepts for games where the grand coalition is not optimal.} according to $\nu$, a key question of cooperative game theory is how to distribute the value of the coalition among its members (i.e., to give each player their payoff from joining the coalition).

\paragraph{The Shapley Value:} An important solution concept was proposed by \citetext{Shapley:1953}, based on the concept of marginal contribution. The Shapley value of player $i$, denoted by $SV_{i}(\nu)$, is then equal to a weighted average of the marginal contributions of $i$ to every coalition that she or he can belong to. In more detail, let $\Pi(I)$ be the set of all permutations of the set of $n$ players, $I$, i.e., every $\pi\in\Pi(I)$ is a bijection from $\{1, 2, \ldots , n\}$ to itself, where $\pi(i)$ is the position of player $i$. Let $C_\pi(i)$ be the coalition consisting of the players that precede $i$ in $\pi$. That is, $C_\pi(i) = \{j \in I: \pi(j) < \pi(i)\}$. Then, the Shapley value of $i$ is the average marginal contribution of player $i$ to $C_\pi(i)$ over all permutations $\pi$ of $I$. Formally,

\begin{equation}
SV_i(\nu) = \frac{1}{n!} \sum_{\pi \in \Pi(I)} [\nu(C_\pi(i) \cup \{i\}) - \nu(C_\pi(i))].
\label{SV1}
\end{equation}

Equivalently, we can write the Shapley value in a (computationally) simpler form:
\begin{equation}
\label{SV2}
SV_i(\nu) = \sum_{C \subseteq I \setminus \{i\}} \frac{|C|! (n - |C| - 1)!}{n!} [\nu(C \cup \{i\}) - \nu(C)].
\end{equation}

The Shapley value may also be expressed in terms of the Harsanyi Dividends and synergies that we discussed earlier in this section. Intuitively, the Harsanyi dividend of a coalition represents the unique synergy achieved by the coalition. Since this synergy is achieved jointly by the coalition members, then it is only ``fair'' that it is distributed evenly among them. In formal terms:
\begin{equation}
\label{SV3}
SV_i(\nu) = \sum_{C \in \{C \midd C \subseteq I \text{ and } i \in C\}} \frac{\Delta_{\nu}(C)}{|C|}.
\end{equation}
We refer the reader to the work of \citetext{Pruzhansky:2005} and \citetext{Macho-Stadler:et:al:2010} for an in-depth introduction to Harsanyi dividends and their relation to solution concepts in cooperative game theory.

Among the many interesting properties of the Shapley value is the fact that it is the unique solution concept satisfying the following four desirable axioms:

\begin{itemize}
	\item \textbf{Efficiency}---the whole value of the grand coalition is distributed. That is, $\sum_{i \in I} SV_{i}(\nu) = \nu(I)$;
	\item \textbf{Symmetry}---the payoffs of the players do not depend on their identities;
	\item \textbf{Null-player}---players whose marginal contributions are equal to zero for all coalitions receive zero payoff; and
	\item \textbf{Additivity}---for any two games with the same player set, $(I, \nu_1)$ and $(I, \nu_2)$, the sum of the payoffs for both games is equal to the payoff of the sum of the games for all players. That is, for every player $i$ in $I$, $SV_{i}(\nu_1) + SV_{i}(\nu_2) = SV_{i}(\nu_1 + \nu_2)$, where $(\nu_1 + \nu_2)(C) = \nu_1(C) + \nu_2(C)$.
\end{itemize}

\citetext{Young:1985} showed that the Null-Player and Additivity axioms can be replaced by the strong monotonicity axiom:

\begin{itemize}
	\item \textbf{Strong Monotonicity}---for any two games $\nu$ and $\omega$ with the same player set $I$, if for a player $i$ we have $\MC_{\nu}(S,i) \geq \MC_{\omega}(S,i)$ for every coalition $S\subseteq (I \setminus \{i\})$, then $SV_i(\nu) \geq SV_i(\omega)$.
\end{itemize}

In the network theory context, this axiom can be interpreted as follows. If we consider two graphs with the same node set, then node $i$ will be worth more in the graph in which it has a better (i.e., more \textit{central}) position. Note that there exist in the literature many other characterisations of the Shapley value, i.e., properties or axioms that uniquely define it (the set of axioms above is just one example).

\paragraph{Semivalues:} The Shapley value is an example of a more general class of division schemes called semivalues. The Banzhaf index of power \cite{Banzhaf:1965}, which was originally introduced for voting games, is another example. In particular, semivalues assign to each player the expected value of his or her marginal contribution to any coalition with some probability distribution over the size of the coalition. In the context of semivalues, one can interpret the Shapley value as the expected marginal contribution of a player $i$ such that the probability of $i$ joining a random coalition of a certain size is equal to the probability of $i$ joining a random coalition of any other size. The Banzhaf index, on the other hand, can be interpreted as the expected marginal contribution such that a player joins any random coalition (regardless of size) with uniform probability. In general, semivalues allow for any distribution over the size of coalitions, but the probability of drawing any two coalitions of the same size must be the same. We will denote by $\beta(k)$ the probability that a player joins a random coalition of size $k$. Formally, the general formula for semivalues is as follows:
\begin{equation}\label{SEMI}
\phi^\beta_i(\nu)  = \sum_{0 \leq k \leq n-1} \beta(k) \mathbb{E}[\MC_{\nu}(C^k(I \setminus \{i\}),i)],
\end{equation}
where $C^k(I \setminus \{i\})$ is a random variable ranging over the set of all coalitions of size $k$ being drawn with uniform probability from the set $I \setminus \{i\}$ and $\mathbb{E}[\cdot]$ is the expected value operator. Alternatively, we present a formulation of semivalues without the need for an expected value operator below:

\begin{equation}
\phi_i(\nu) = \sum_{k = 0}^{n-1} \sum_{C \in C^k(I \setminus \{i\})} \beta(k) \frac{MC_\nu(i, C)}{\binom{n-1}{k}},
\end{equation}
where we slightly abuse notation for $C^k(I \setminus \{i\})$ now referring to all subsets of size $k$ of the set $I \setminus \{i\}$. The Shapley value~\cite{Shapley:1953} and the \emph{Banzhaf index} of
	power~\cite{Banzhaf:1965} are defined by the $\beta$-functions $\beta^{\mathit{Shapley}}$ and $\beta^{\mathit{Banzhaf}}$, respectively, where:
	\begin{equation}\nonumber
	\beta^{\mathit{Shapley}}(k)=
	\frac{1}{n} ~~\mbox{ and }~~
	\beta^{\mathit{Banzhaf}}(k)=\frac{\binom{n-1}{k}}{2^{n-1}}.
	\end{equation}

It is worth noting that the Banzhaf index is not an \textit{efficient} semivalue, i.e., it does not necessarily distribute the entire worth of the grand coalition among the players. In this sense, the Banzhaf index is not suitable as a payoff-division scheme.\footnote{\footnotesize It is possible to rescale the Banzhaf index, however it then loses the property of \textit{Additivity}.} Interestingly, the Shapley value is in fact the only efficient semivalue, i.e., the only one satisfying the aforementioned Efficiency property.

The significance of semivalues in the context of game-theoretic centralities is that they provide a method for specifying the importance of certain types of synergy. A trivial example is to give a weight of $1$ to the empty coalition (i.e., $\beta(0) = 1$) and $0$ to all other coalitions. In this case, synergy is not considered at all. If this semivalue is applied to group degree, group betweenness or group closeness, then the resulting measure is the standard degree, betweenness or closeness measure. In essence, this answers the question ``what can a player achieve on his or her own.'' Going further, by varying the ratio between the weight of the empty coalition and the non-empty coalitions, it is possible to specify the relative impact of synergy on the ranking of players. On the other hand, giving coalitions of size $n-1$ a weight of $1$ (i.e., $\beta(n-1) = 1$) and all other coalitions a weight of $0$, the resulting semivalue answers the question ``how much would all the players lose were a single player to leave.'' Semivalues give the possibility of fine-tuning the types of contributions and synergies that are important in determining the worth of players or---in the case of networks---nodes.

\paragraph{Owen Value:} Importantly, semivalues assume that all types of interactions between players are possible (i.e., anyone can contribute to any coalition).
To relax this assumption, \cite{Owen:1977} introduced a solution concept---now known as the Owen value---that limits interactions of players to only the other players in their communities. This requires the extension of the standard cooperative setting to include a community/coalition structure $\CS = \{Q_1, Q_2, \ldots Q_m\}$, which is simply a non-overlapping partition of the player set $I$. The Owen value divides the payoff of any \textit{a priori} coalition structure $\CS$ among the communities in $\CS$ and the value of each of these communities is distributed fairly and efficiently among its members. Assuming $i \in Q \in \CS$, formally, the Owen Value is defined as follows:

\begin{align}
\textit{OV}_i(\nu, \CS) = \sum_{T \in \CS \setminus Q} \sum_{C \in Q} \frac{1}{|\CS|\binom{|\CS| - 1}{|T|}} \frac{1}{|Q|\binom{|Q| - 1}{|C|}} \MC\Big( (\bigcup T) \cup C \Big)
\end{align}

Now, when $\CS = \{I\}$ or $\CS = \{\{i\}\}_{i \in I}$, the Owen value is equivalent to the Shapley value. As such, the Owen value is a generalization of the Shapley value; one that does not generalise the $\beta$ function (as semivalues do), but rather generalises the assumed coalition structure $\CS$. While the details are beyond the scope of this article, we will mention that the Owen value is uniquely characterised by the following axioms: Efficiency, Null Player, Symmetry, Linearity, and Coalitional Symmetry. We have already introduced the first three of these axioms when discussing the Shapley value. If a payoff division scheme, $\phi_i(\nu)$, satisfies linearity, then for any $\lambda \in \mathbb{R}$ and cooperative games $\nu_1$ and $\nu_2$, we have $\phi_i(\lambda \nu_1 + \nu_2) = \lambda\phi_i(\nu_1) + \phi_i(\nu_2)$. Coalitional Symmetry implies that if two communities in $\CS$ contribute the same value to all coalitions of communities, then their payoffs (i.e., the sum of the payoffs of their members) are the same.

%

\paragraph{Coalitional Semivalues:} Another step in this line of research was taken by \citeauthor{Szczepanski:et:al:2014} 
\cite{Szczepanski:et:al:2014}, who proposed a generalisation combining both the Owen value and semivalues; they called it \emph{coalitional semivalues}. Formally, given a coalition structure, $\CS$, and discrete probability distributions: $\beta:\{0, \dots, |\CS|-1\}\to[0,1]$ and $\alpha_Q:\{0, \dots, |Q|-1\}\to[0,1]$ coalitional semivalues are defined by:
\begin{align}
\gamma_i(\nu, \CS) = \sum_{k=0}^{|\CS|-1} \beta(k) \sum_{l=0}^{|Q|-1}  \alpha_Q(l) \mathbb{E} \left[\MC\Big(\big(\bigcup T^k(CS \setminus \{Q\})\big) \cup C^l(Q \setminus \{i\}), i\Big)\right]
\end{align}
where $Q$ is the coalition in $\CS$ that player $i$ belongs to, $T^k(\CS \setminus \{Q\})$ is a random variable over subsets of size $k$ chosen from $\CS \setminus \{Q\}$ with uniform probability, $C^l(Q \setminus \{i\})$ is a random variable over subsets of size $l$ chosen from $Q \setminus \{i\}$ with uniform probability, and $\mathbb{E}$ is the expected value operator. 
%
%
The coalitional semivalue is equivalent to the Owen value when $\beta(k) = \frac{1}{|\CS|}$ and $\alpha_Q(l) = \frac{1}{|Q|}$.

\paragraph{Solution Concepts for Games with Overlapping Communities:} None of the solution concepts discussed thus far considers overlapping communities. To address this issue, \citeauthor{Albizuri:2006} \cite{Albizuri:2006} generalised the Owen value to situations where the \emph{a priori} coalition structure $\CS$ contains overlapping communities; they called this generalisation the \textit{Configuration value}. Formally, it is defined as follows, where $\mathcal{T}(i) = \{Q: Q \in \CS \text{ and } i \in Q\}$:
-------------------------
%
\begin{align}
&\chi_i(\nu, \CS) = \sum_{\substack{T \subseteq \CS \\ T \cap \mathcal{T}(i) = \emptyset}} \sum_{Q \in \mathcal{T}(i)} \sum_{\substack{C \subseteq Q \\ i \not\in C}} \lambda \MC\Big(\big(\bigcup T\big) \cup C, i\Big),
\end{align}
where
\begin{align*}
\lambda = \frac{|T|!(|\CS| - |T| - 1)!}{|\CS|!} \frac{|C|! (|Q| - |C| - 1)!}{|Q|!}
\end{align*}

\paragraph{Interaction Indices:} 
The concept of Interaction indices is intimately connected to payoff division schemes and synergies. We already mentioned that synergies can be used to evaluate the similarity of nodes, since if two players have negative synergy, then this is likely to indicate that their functions are similar and therefore redundant. Given this, \citetext{Owen:1972} defined the interaction between two players as their expected synergy given a random coalition. We refer to this index as the Shapley value interaction index, which is defined as follows:

\begin{align}
I^{\mathit{Shapley}}_{i,j}(\nu) = \sum_{\pi \in \Pi(I^{i \wedge j}) } \frac{S_\nu(C_\pi(\{i,j\}),i,j)}{(n - 1)!},
\label{equation:Shapley_interaction_index}
\end{align}

where $I^{i \wedge j}$ is the set $I$ such that the elements $i \in I$ and $j \in I$ are replaced by a single element $\{i,j\}$. In other words, this can be interpreted as the Shapley value of the player $\{i,j\}$ in the game where $i$ and $j$ join to create one player minus the Shapley value of player $i$ in the game where $j$ is removed minus the Shapley value of player $j$ in the game where $i$ is removed. \citetext{Grabisch:1999} continued this work and introduced the Banzhaf interaction index. \citetext{Szczepanski:2015b} generalised these concepts further by introducing semivalue interaction indices, defined below:

\begin{equation}
I^{\mathit{SEMI}}_{i,j}(\nu) = \sum_{k = 0}^{n-2} \sum_{C \in C^k(I \setminus \{i,j\})} \beta(k) \frac{S_\nu(C,i,j)}{\binom{n - 2}{k}}
\label{equation:semivalue_interaction_index}
\end{equation}


\citetext{Szczepanski:2016b} extended the concept further to allow for a priori community structures and defined the Coalitional semivalue interaction indices. Assuming $i$ belongs to community $P$, and $j$ to $Q$, we define them as follows:

\begin{align}
&I_{i,j}^{\mathit{CoSemi}}(\nu) =\sum_{k=0}^{|\CS|-2} \beta(k) \sum_{l = 0}^{|Q|+|P|- 2}  \alpha_{|P|+|Q|-2}(l) \nonumber\\ &\sum_{T \in C^k(\CS \setminus \{P,Q\})} \sum_{C \in C^l(N \setminus \{i,j\})} \frac{S_{\nu} \bigg( \bigcup T \cup C, i, j \bigg)}{\binom{|\CS|-2}{k}\binom{|P|+|Q|-2}{l}}\text{,}
\end{align}
\noindent where $C^k(P) = \{C: C \subseteq P \text{ and }  |C| = k\}$ are subsets of size $k$ of the set $P$.



We finish our discussion of concepts from cooperative game theory by introducing a generalisation of characteristic functions, where the order of group of players impacts their value.

\paragraph{Generalised cooperative games:} Of relevance to some of the work related to this article are generalised cooperative games. Like its standard counterpart, a generalised cooperative game also contains a set of $n$ players, $I = \{1, 2, \ldots , n\}$. However the function that defines the game, which is called a \emph{generalised characteristic function}, is slightly different. Generalised characteristic functions take into account the \textit{order} of players. Consequently, the value of an \textit{ordered coalition} not only depends on the identities of the players therein, but also their order. Formally, a generalised characteristic function is a function from the set of all ordered subsets of $I$ to the real numbers. We will denote by $\pi_C: C \mapsto \{1, 2, \ldots, |C|\}$ a permutation of the set $C$ 
and by $\Omega(I)$ the set of all ordered coalitions of $I$, i.e., $\Omega(I) = \bigcup_{C \subseteq I} \Pi(C)$. A generalised characteristic function, then, is denoted as follows: $\nu^*: \Omega(I) \rightarrow \mathbb{R}$.

\paragraph{Unanimity Basis and Dividends for generalised Cooperative Games:} The unanimity basis for generalised cooperative games is defined in a similar fashion to the unanimity basis for regular cooperative games. First, we must define a partial order, $\widetilde{\subseteq}$, on the set of ordered coalitions $\Omega(I)$. For $S,C \in I$ we will define $\widetilde{\subseteq}$ as follows:
\[
\pi_S \widetilde{\subseteq} \pi_C \text{ if and only if } S \subseteq C \text{ and } \forall_{s,t \in S} \pi_C(s) < \pi_C(t) \text{ implies } \pi_S(s) < \pi_S(t)
\]
If $S \subset C$, i.e., the inclusion is strict, then we will write $\pi_S \widetilde{\subset} \pi_C$. The generalised unanimity basis, then, is the set of all generalised characteristic functions indexed by ordered coalitions of the form:
\[
w^*_{\pi_C}(\pi_S) = \begin{cases}
1 & \text{if } \pi_C \widetilde{\subseteq} \pi_S\\
0 & \text{otherwise}
\end{cases}
\]
In other words, any generalised characteristic function $\nu^*$ can be written in the form:
\[
\nu^*(\pi_C) = \sum_{\pi \in \Omega(I)} \Delta^*_{\nu^*}(\pi) w^*_{\pi}(\pi_C),
\]
where $\Delta^*_{\nu^*}(\pi), \forall \pi\in\Omega(I)$ are coefficients, which we call generalised dividends. \citetext{Sanchez:Bergantinos:1997} showed that:
\begin{equation}\label{eqn:deltaStar}
\Delta^*_{\nu^*}(\pi_C) = \nu^*(\pi_C) - \sum_{\pi \widetilde{\subset} \pi_C} \Delta^*_{\nu^*}(\pi).
\end{equation}

Now if, for all ordered coalitions $\pi_1, \pi_2$ with the same set of players $S$ (i.e., $\pi_1, \pi_2 \in \Pi(S)$) we have a generalised cooperative game such that $\nu^*(\pi_1) = \nu^*(\pi_2)$, then this game is just a regular cooperative game. We can rewrite the basis for this game as $u^*_S = \sum_{\pi_S \in \Pi(S)} w_{\pi_S}$, which is simply equal to the unanimity basis of a regular cooperative game with Harsanyi dividends (with the difference, that it accepts ordered coalitions, rather than unordered ones).

\subsection{Graph-Theoretic Concepts}

A \emph{network} is a \emph{weighted digraph}~$D=(V, E,\omega)$, where~$V$ is a set of nodes,~$E$ is a set of edges, i.e., ordered pairs~$(v,u)$ of nodes in~$V$ with $v\neq u$, and $\omega: E \rightarrow \mathbb{R}^+$ is a weight function from edges to the positive real numbers. A weighted digraph $D=(V,E,\omega)$ is said to be undirected if (i) $(v,u)\in E$ implies $(u,v)\in E$ and (ii) $\omega((v,u))=\omega((u,v))$ for all $(u,v)\in E$. In that case,~$D$
is also said to be a weighed graph. A weighted graph or digraph $D=(V,E,\omega)$ is said to be \textit{unweighted} if $\omega(e)=1$ for all $e\in E$. Unweighted graphs and unweighed digraphs we also refer to as simply graphs and digraphs, respectively, and, in that case, we will generally omit the reference to~$\omega$ in the signature.


A \textit{path}~$\pi_{st}$ from a source node~$s$ to destination node~$t$ in a graph~$G$ is an ordered set $(v_0, v_1, \ldots, v_k)$ such that~$v_0 = s$ and~$v_k = t$ and $(v_i, v_{i+1}) \in E$ for all~$i$ with $1\le i<k$. We will define $\mathit{Paths}(s,t)$ as the set of paths between nodes $s$ and $t$ and
$\mathit{SP}(G)$ as the set of all shortest paths in $G$. We define the set of \emph{neighbours} of a node~$v$ by $E(i)=\set{j\midd (i,j)\in E}$. The neighbour set of a subset~$C$ of nodes is defined as $E(C)=\bigcup_{i\in C}E(i)\setminus C$.
We refer to the \emph{degree} of a node~$v$ by $\mathit{deg}(v) = |E(v)|$. The \emph{distance} from a node~$s$ to a node~$t$ is denoted by~$\mathit{dist}(s, t)$ and is defined as the length  of the shortest path (i.e., sum of the edge weights between subsequent nodes in the path) between~$s$ and~$t$. The distance between a node~$v$ and a subset of nodes~$C \subseteq V$ is denoted by $\mathit{dist}(C, v) = \min_{u \in C} \textit{dist}(u, v)$. Paths, neighbours, and distances are prominent in definitions of \emph{network centrality measures}, i.e., functions that associate with each node a real value that represents its centrality, and we will refer to them as \emph{(graph) items.}

By $G[S]$, where $G=(V,E)$ and $S \subseteq V$, we denote the subgraph of $G$ induced by $S$. In other words, $G[S] = (S, E')$, where $E = \set{(u,v) \midd (u,v) \in E \text{ and } u \in S \text{ and } v \in S}$.
We say that a graph is \textit{connected} if and only if there exists a path between any two nodes in the graph. We define the \textit{connected components} of a graph $G(V,E)$, denoted by $\mathcal{K}_G(V)$, as the set of sets of nodes that represent all maximal sub-graphs of $G$ such that they are connected. In other words, $C \in \mathcal{K}_G(V)$ if and only if $C$ is connected, and for any nonempty $C' \subseteq V\setminus C$, $C \cup C'$ is not connected. For $C \subseteq V$ we will denote by $\mathcal{K}_G(C)$ the maximal connected components of $G[C]$ (or just $\mathcal{K}(C)$ when $G$ is apparent from the context).

\begin{figure}[thbp]
	\center\includegraphics[width=10.5cm]{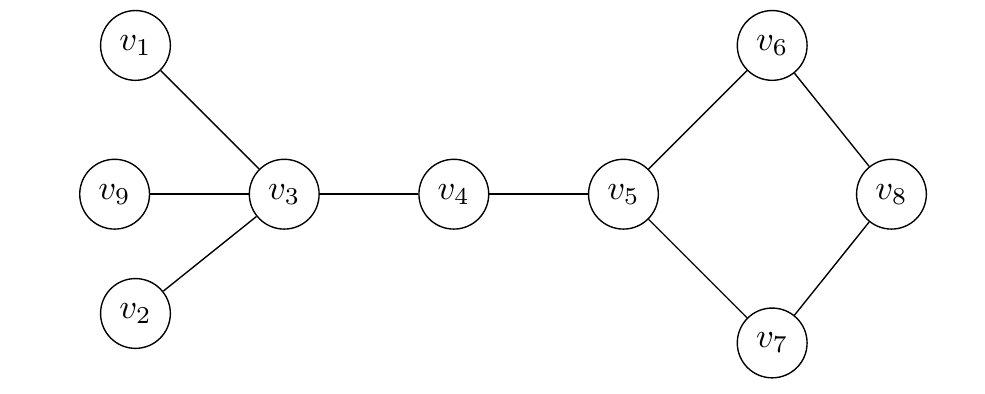}
	\caption{Sample network of $8$ nodes.}
	\label{fig:Classical:Centrality:Example}
\end{figure}

\begin{example}
	Figure~\ref{fig:Classical:Centrality:Example} presents a sample graph of $9$ nodes, which we will refer to as $G$. Since $G$ is connected, then $\mathcal{K}(V) = \{ \{v_1, v_2, v_3, v_4, v_5, v_6, v_7, v_8, v_9\} \}$. Let us consider a disconnected group of nodes, $C = \{v_1, v_2, v_3, v_6, v_7, v_8 \}$. The maximal subsets of $C$ that are connected in $G$ (i.e., the connected components of $C$), are given by: $\mathcal{K}(C) = \{\{ v_1, v_2, v_3\}, \{ v_6, v_7, v_8 \} \}$.
\end{example}

The adjacency matrix of a graph $G$, denoted by $A_G$, is the matrix indexed by the set of nodes $V$ in $G(V,E)$ such that $A_G[i,j] = 1$ if $(i,j) \in E$ and $A_G[i,j] = 0$ otherwise.

\paragraph{Classic Node Centrality:}
We now present the three classic notions of node centrality as introduced by \citeauthor{Freeman:1979} \cite{Freeman:1979} and Eigenvector centrality due to \citetext{Bonacich:1972}.

\begin{enumerate}
	\item \textit{Degree Centrality} ranks nodes according to how many neighbours they have. The larger the neighbourhood of a node, the more central it is. Formally:
	\[\mathit{degree}(v) = |E(v)|\]
	
	\begin{example}
		In Figure~\ref{fig:Classical:Centrality:Example}, the node with the highest degree centrality is $v_3$, having a degree of $4$ (its neighbors are $v_1, v_2, v_4, $ and $v_9$). Next, $v_5$ has 3 neighbors ($v_4, v_6,$ and $v_7$). The third place in the ranking is a four-way tie between the nodes $v_4$, $v_6$, $v_7$, and $v_8$ each of which have degree $2$. Finally, nodes $v_1$, $v_9$ and $v_2$ are ranked last.
	\end{example}
	
	\item \textit{Betweenness Centrality} ranks a node according to how many shortest paths in the network it lies on. Often, the additional constraint is added that paths do not contribute to the rank of their source and destination nodes. Formally, let $\sigma_{(s,t)}$ be the number of shortest paths between $s$ and $t$, and $\sigma_{(s,t)}(v)$ be the number of shortest paths between $s$ and $t$ that visit $v$. 
	Given this, betweenness centrality is defined as follows:
	\[\mathit{betweenness}(v) = \sum_{s,t \in V \setminus \{v\}}\frac{\sigma_{(s,t)}(v)}{\sigma_{(s,t)}}\]
	A simpler version of betweenness is called stress centrality. As opposed to betweenness, which normalises the paths that a node lies on by the number of paths between any source and target node, stress centrality simply counts the number of paths that a node lies on. It is defined as follows:
	\[\mathit{stress}(v) = \sum_{s,t \in V \setminus \{v\}} \sigma_{(s,t)}(v)\]
	
	\begin{example}
		In Figure~\ref{fig:Classical:Centrality:Example}, the node with the highest betwenness centrality is $v_3$, with a centrality of $36$. Nodes $v_4$ and $v_5$ follow with a centrality of $33$ and $32$, respectively. Continuing, nodes $v_6$ and $v_7$ have a centrality of $6$ each. Finally, $v_8$ has a centrality of $1$ and all the other nodes have a centrality of $0$, since they do not play any intermediary role in any of the paths in the network.
	\end{example}
	
	\item \textit{Closeness Centrality} ranks nodes based on their distances to other nodes. The most classical version of the centrality ranks a node according to the sum of the distances from it to any other node in the network:
	\[\mathit{closeness}(v) = \sum_{u \in V} \mathit{dist}(v,u).\]
	This results in an inverse ranking (i.e., nodes that are more important have smaller value). A popular variation on closeness \cite{Michalak:et:al:2013b}, which results in a ranking such that more important nodes have higher value, uses some non-increasing function of distance, $f: \mathbb{R} \rightarrow \mathbb{R}$, rather than distance itself. This generalised form of closeness is referred to as the influence game by \citetext{Michalak:et:al:2013b}. For clarity, however, we refer to it simply as generalised closeness throughout this article. Formally, generalised closeness is defined as follows:
	\[\mathit{generalised\_closeness}(v) = \sum_{u \in V} f(\mathit{dist}(v,u)).\]
	When $f(k) = \frac{1}{k}$ the resulting measure is called \textit{harmonic closeness} \cite{Boldi:Vigna:2013}.

	\begin{example}
		In Figure~\ref{fig:Classical:Centrality:Example}, the node with the best closeness centrality is $v_4$, since the sum of its distances to other nodes is $15$. Next, for nodes $v_3$ and $v_5$ this value is $16$, for nodes $v_6$ and $v_7$ it is $21$, for $v_1$, $v_2$ and $v_9$ it is $23$, and node $v_8$ comes in last with $26$.
	\end{example}

	\item \textit{Eigenvector Centrality}, $C_E(v)$, assigns to each node $v$ in $G$ the $v^{\text{th}}$ entry of an eigenvector with the largest eigenvalue of the adjacency matrix of $G$. More formally, let $\vec{x}$ be an eigenvector with the largest eigenvalue, $\lambda$. That is, if $A_G$ is the adjacency matrix of $G$, then $\lambda$ is the largest value such that there exists a vector $\vec{y}$ for which we have $A_G \vec{y} = \lambda \vec{y}$. Moreover, let $\vec{x}$ be an eigenvector for this value, i.e., $A_G \vec{x} = \lambda \vec{x}$. Then:
		\[
		\textit{eigenvector}(v) = \vec{x}[v] 
		\]
	\begin{example}
		In Figure~\ref{fig:Classical:Centrality:Example}, the node with the highest eigenvector centrality is $v_4$, with a centrality of $0.486919$. Next is $v_5$ with $0.401595$, then $v_6$ and $v_7$ with $0.395147$, $v_3$ with $0.350707$, $v_8$ with $0.255378$, and $v_1$, $v_2$ and $v_9$ with $0.210938$.
	\end{example}

\end{enumerate}

\paragraph{Group Centrality:} For the purposes of this article, we define group centrality as follows:

\begin{definition}[Group centrality]
	In general, we will use the variable $\psi$ to denote a group centrality measure. Formally, $\psi$ maps any graph~$G=(V,E)$ onto a characteristic function $\nu: 2^V \rightarrow \mathbb{R}$. 
\end{definition}

We present below some classic group centralities due to \citeauthor{Everett:Borgatti:1999} \cite{Everett:Borgatti:1999}. 
These group centrality measures are a natural extension of the centrality measures introduced by \citeauthor{Freeman:1979} \cite{Freeman:1979}. 

\begin{enumerate}
	\item \textit{Group Degree Centrality} of a group of nodes $S$ is defined as the size of the neighbourhood of this group. 
	Formally, group degree centrality in a graph $G$ is defined as: \[\psi^D(G)(S) = \nu_G^D(S) = |\{v: v \in E(S) \setminus S\}|;\] 
	
	\item \textit{Group Betweenness Centrality} of a group of nodes $S$ is defined as the number of shortest paths that visit at least one node in $S$. Often (but not always), the additional assumption is made that the source and destination nodes of these paths cannot belong to $S$. 
	Let $\sigma_{(s,t)}$ be the number of shortest paths between the nodes $s$ and $t$, and~$\sigma_{(s,t)}(S)$ be the number of shortest paths between $s$ and $t$ that visit at least one node in~$S$. 
	Formally, group betweenness centrality in a graph $G$ is defined as: \[\psi^B(G)(S) = \nu_G^B(S) = \sum_{s,t \in V \setminus S} \frac{\sigma_{(s,t)}(S)}{\sigma_{(s,t)}}.\]
	
	\item \textit{Group Closeness Centrality} of a group of nodes~$S$ is defined as the sum of the distances from~$S$, to any node outside of~$S$. This results in a ranking, where groups with lower value are---on average---closer to the nodes outside of the group, and therefore more central. Formally, group closeness centrality in a graph $G$ is defined as: \[\psi^{CL}(G)(S) = \nu_G^{\mathit{CL}}(S) = \sum_{v \in V \setminus S} dist(S,v);\]
	Similarly to closeness centrality, group closeness centrality can be generalised such that more central groups have a higher rank. This extension was called the \textit{influence game} by \citetext{Michalak:et:al:2013b}, and it is defined with the help of a  non-increasing function $f: \mathbb{R} \rightarrow \mathbb{R}$ in the following manner:
	\begin{equation}
	\psi^{CL}_f(G)(S) = \nu^{CL}_{G,f}(S) = \sum_{v \in V \setminus S} f(dist(S,v));
	\label{equation:general_group_closeness}
	\end{equation}
	When $f(k) = \frac{1}{k}$, the resulting measure is called \textit{harmonic group closeness} \cite{Boldi:Vigna:2013}.
\end{enumerate}
In all instances, when the graph $G$ in question is obvious in a given context, we skip the subscript $G$ when referring to the characteristic function. For $\psi^{CL}(f,G)(S)$ and $\nu^{CL}_{G,f}(S)$, we skip the subscript $f$ whenever it is obvious in a given context. This allows us to reduce a clutter of notation.

\section{The Definition of Game-Theoretic Centrality Measure}\label{section:definition}
We are now in a position to define game-theoretic centrality, which can be viewed as a method to rank nodes by aggregating the centralities of (typically all) groups of nodes in a network. In particular, in this approach nodes are treated as players in a cooperative game and their groups as coalitions. 
The characteristic function is typically some group centrality measure (such as those described above) and a chosen solution concept is the method of aggregation. In the context of this article, a pair of any given group centrality measure and solution concept can be considered a game-theoretic centrality measure, since it yields a ranking of nodes in any graph.

\begin{definition}[Game-theoretic centrality measure] A \textit{game-theoretic centrality measure} can be defined as a pair~$(\psi, \phi)$ consisting of a group centrality $\psi$ and a cooperative solution concept~$\phi$. Givens $G = (V,E)$, we will sometimes use a characteristic function $\nu_G: 2^V \rightarrow \mathbb{R}$ to refer to the group centrality measure.
\end{definition}



Let us now look at the following example illustrating the concept of game-theoretic centrality.

\begin{figure}
	\[
	\begin{tikzpicture}[auto]
	\tikzstyle{every node}=[fill=white,circle,draw,minimum size=1.5em,inner sep=0pt]
	\node (1) at (0,0) {$a$};
	\node (2) at (2, 1.5) {$b$};
	\node (3) at (4.25, 1.5) {$c$};
	\node (4) at (2,-1.5) {$d$};
	\node (5) at (4.25,-1.5) {$e$};
	
	\draw (1) -- (2) -- (3);
	\draw (2)	-- (4);
	\draw (1) -- (4) -- (5);
	\end{tikzpicture}
	\]
	\caption{A simple example.}
	\label{figure:betweenness_bph}
\end{figure}
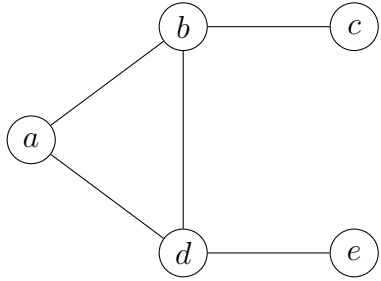

\begin{example}\label{example:betweenness_bph}
	Consider betweenness centrality and group betweenness centrality in the network~$G$ depicted in Figure~\ref{figure:betweenness_bph}.
	Observe that, besides the direct edges, there are also the following shortest paths between nodes	:
	\[
	abc, ade, bde, cba, cbd, cbde, dbc, eda, edb, edbc\text.
	\]
	It can then easily be established that the betweenness centrality for nodes~$a$, $c$, and~$e$ is~$0$, whereas that of~$b$ and~$d$ totals to~$6$.  
	The group centralities $\nu^B(C)$ (omitting the reference to~$G$) of the groups of nodes (coalitions)~$C$ are summarised in Table~\ref{table:betweenness_bph}.
	We can now use  the Shapley value to aggregate the group centralities for the individual nodes. 
	In this way, we find that the game-theoretical betweenness 
	centralities for the Shapley value of the nodes~$a$,~$b$,~$c$,~$d$, and~$e$ are, respectively, $\frac{-2}{5}$, $\frac{3}{10}$, $\frac{-7}{6}$, $\frac{3}{10}$, and $\frac{-7}{6}$. 
\end{example}

\begin{table}
	\small
	\[
	\begin{array}{cc@{\qquad}cc@{\qquad}cc@{\qquad}cc}
	\toprule
	C	&	\nu^B(C)	& C	&	\nu^B(C)	&C	&	\nu^B(C)	&C	&	\nu^B(C)	\\\midrule
	\emptyset	
	& 0	&	\set{a,b}	&	4	&	\set{a,b,c}	&	0	&	\set{a,b,c,d}	&	0	\\
	\set{a}	 	& 0	&	\set{a,c}	&	0	&	\set{a,b,d}	&	2	&	\set{a,b,c,e}	&	0	\\
	\set{b}	 	& 6	&	\set{a,d}	&	4	&	\set{a,b,e}	&	2	&	\set{a,b,d,e}	&	0	\\
	\set{c}	 	& 0	&	\set{a,e}	&	0	&	\set{a,c,d}	&	2	&	\set{a,c,d,e}	&	0	\\
	\set{d}	 	& 6	&	\set{b,c}	&	0	&	\set{a,c,e}	&	0	&	\set{b,c,d,e}	&	0	\\
	\set{e}	 	& 0	&	\set{b,d}	&	6	&	\set{a,d,e}	&	0	&	\set{a,b,c,d,e}	&	0	\\
	&	&	\set{b,e}	&	4	&	\set{b,c,d}	&	2	&	 				&	\\
	&	&	\set{c,d}	&	4	&	\set{b,c,e}	&	0	&	 				&	\\
	&	&	\set{c,e}	&	0	&	\set{b,d,e}	&	2	&	 				&	\\
	&	&	\set{d,e}	&	0	&	\set{c,d,e}	&	0	&	 				&	\\\bottomrule										
	\end{array}
	\] 
	\caption{The betweenness centralities of the coalitions of the network in Figure~\ref{figure:betweenness_bph}.}
	\label{table:betweenness_bph}
\end{table}

Let us now discuss the differences between the three approaches to centrality that we mentioned---classic, group and game-theoretic---based on a sample network.

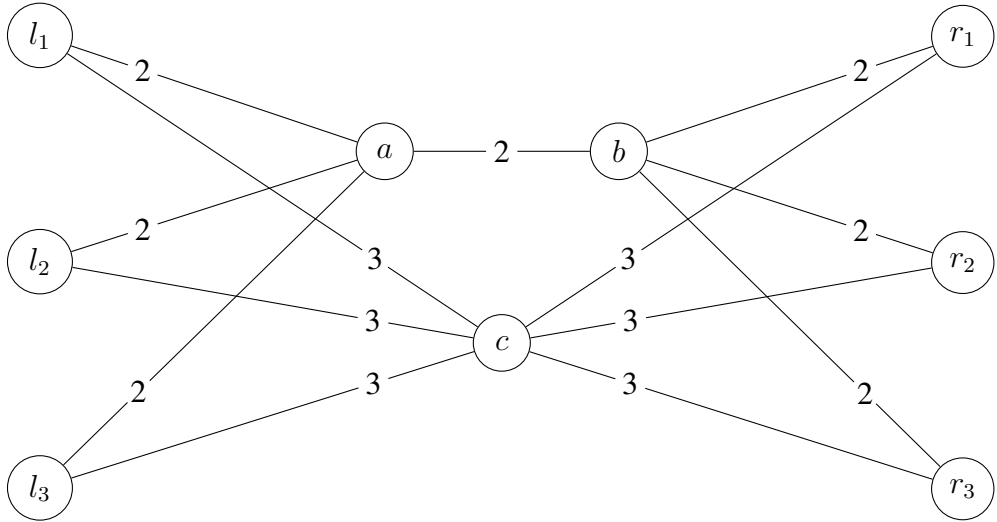
\begin{figure}
	\centering
	\begin{tikzpicture}[scale=.6]
	\tikzset{edge/.style = {->,> = latex'}}
	\node[draw=black,circle,minimum size=.75cm] (3)  {$c$};
	\node[draw=black,circle,minimum size=.75cm] (1) [above left = 2cm and 1cm of 3] {$a$};
	\node[draw=black,circle,minimum size=.75cm] (2) [above right = 2cm and 1cm of 3] {$b$};
	\node[draw=black,circle,minimum size=.75cm] (4) [above left = 3.5cm and 5.5cm of 3] {$l_1$};
	\node[draw=black,circle,minimum size=.75cm] (5) [above left = 0.5cm and 5.5cm of 3] {$l_2$};
	\node[draw=black,circle,minimum size=.75cm] (6) [above left = -2.5cm and 5.5cm of 3] {$l_3$};
	\node[draw=black,circle,minimum size=.75cm] (7) [above right = 3.5cm and 5.5cm of 3] {$r_1$};
	\node[draw=black,circle,minimum size=.75cm] (8) [above right = 0.5cm and 5.5cm of 3] {$r_2$};
	\node[draw=black,circle,minimum size=.75cm] (9) [above right = -2.5cm and 5.5cm of 3] {$r_3$};
	
	\draw[-] (1) -- node[draw=white,white,fill,circle]{} node[draw=none]{2} (2);
	
	\draw[-] (1) -- node[draw=white,near end,white,fill,circle]{} node[near end,draw=none]{2} (4);
	\draw[-] (1) -- node[draw=white,near end,white,fill,circle]{} node[near end,draw=none]{2} (5);
	\draw[-] (1) -- node[draw=white,near end,white,fill,circle]{} node[near end,draw=none]{2} (6);
	\draw[-] (2) -- node[draw=white,near end,white,fill,circle]{} node[near end,draw=none]{2} (7);
	\draw[-] (2) -- node[draw=white,near end,white,fill,circle]{} node[near end,draw=none]{2} (8);
	\draw[-] (2) -- node[draw=white,near end,white,fill,circle]{} node[near end,draw=none]{2} (9);
	\draw[-] (3) -- node[draw=white,near start,white,fill,circle]{} node[near start,draw=none]{3} (4);
	\draw[-] (3) -- node[draw=white,near start,white,fill,circle]{} node[near start,draw=none]{3} (5);
	\draw[-] (3) -- node[draw=white,near start,white,fill,circle]{} node[near start,draw=none]{3} (6);
	\draw[-] (3) -- node[draw=white,near start,white,fill,circle]{} node[near start,draw=none]{3} (7);
	\draw[-] (3) -- node[draw=white,near start,white,fill,circle]{} node[near start,draw=none]{3} (8);
	\draw[-] (3) -- node[draw=white,near start,white,fill,circle]{} node[near start,draw=none]{3} (9);
	
	
	
	
	\end{tikzpicture} 
	
	\caption{A graph representing a sample delivery network. Edge weights represent the physical distance between nodes.
	}
	\label{fig:delivery}
\end{figure}

	In particular, let us consider a sample delivery network, where parcels (for instance of data or physical mail) are transferred between nodes. Figure \ref{fig:delivery} presents an example of such a network. Edge weights represent the distance between nodes. 
	Assume that we want to intercept parcels travelling through this network, while the only thing we know is that deliveries are always made via the shortest route. 
	We can try to achieve this by placing a ``mole'' at any of the nodes.
	For this task, we can model the importance of a node $v$ using a variant of betweenness centrality that we call \emph{stress centrality} after \citeauthor{Szczepanski:et:al:2012} \cite{Szczepanski:et:al:2012}:\footnote{Since moles can also affect parcels at their own location, we make the slight modification of also counting towards a node's centrality the shortest paths that originate and end at this node.}
	\[\mathit{stress}(v) = \sum_{s \in V} \sum_{t \in V} \sigma_{(s,t)}(v)\]
	
	If we can only place one mole in this setting, then this ``classic'' approach to betweenness centrality would give us the correct answer: nodes $a$ and $b$ are the best nodes to place the mole, since they control the flow of $44$ different shortest paths each, which is more than any other node. However, what if we can place two moles? Unfortunately, the classic approach to centrality is of little little help here, 
	since placing moles at the two nodes with the highest stress centrality---$a$ and $b$---does not guarantee us the greatest success. We could, for instance, do better by placing a mole on~$b$ and another one on~$c$, as \textit{together} they control a larger number of delivery paths.
	For this analysis, we must consider the group stress centrality, which is defined for any group of nodes $S$ as follows:
	\[\mathit{group\_stress}(S) = \sum_{s \in V} \sum_{t \in V} \sigma_{(s,t)}(S)\]
	We can now see that whereas the group $\{a,b\}$ controls a combined number of $56$ shortest paths, the two optimal groups of size $2$---$\{a,c\}$ and $\{b,c\}$---control $80$ shortest paths each. Although the distinction between the definitions of the node centrality and group centrality in this case may  seem trivial given that only the domain changes between the two definitions, two decades separate the two areas of research.
	
	On the other hand, what would be the best approach to defend against the \textit{random} placement of moles? Assume our objective is to secure as many of the shortest paths as possible, i.e., clearing them from all moles that may be on them. 
	Also assume that we can send a single team to a single node to investigate whether it has been  compromised, and, if so, to neutralise the resident mole. Which node should we send the team to first? This depends strongly on the number of nodes that we believe are compromised. The catch is that by neutralising a mole at a node, we may not succeed in securing any shortest paths through it, as each of them may also be compromised by other moles. Consider the case where we know that two moles are compromised (although any probability distribution that models how many moles are likely to exist in the network can be used).
	
	The question of where to send the team first is a complex issue and it is not immediately obvious what we should optimise for. For instance, there is no ``quickest'' way to ensure that all paths are not compromised. Given that the placement of moles is random, this would require the checking of all nodes. One criterion that we can optimise for is to ensure that we clear as many shortest paths as possible with our first visit. Intuitively, this would mean that we should visit $a$ or $b$ first, since they control the most paths. This, however, is not the case given the following two considerations:
	
	\begin{enumerate}
		\item The shortest paths that $a$ lies on contain more nodes on average than those that $c$ lies on. We note that it is more likely that shortest paths that contain more nodes are compromised, since a mole at any one node on the path would suffice to tamper with deliveries. Given this, if we know that there are two moles, then the likelihood of the second mole lying on a large number of the same shortest paths that $a$ lies on is higher than for $c$. This means that even if we clear $a$ of moles, the likelihood that the shortest paths it lies on are still compromised is higher than for $c$.
		\item Node $a$ lies on many of the same shortest paths as $b$. This means that if $b$ is the other infected node, then clearing $a$ of a mole leaves most of its paths still compromised. However---given that $c$ has few common shortest paths with other nodes---clearing $c$ of a mole will clear most of its shortest paths with absolute certainty.
	\end{enumerate}
	
	As an example, consider all shortest paths from some $l_i$ to some $r_j$ for $i,j \in \set{1, 2, 3}$. There are $9$ paths of the form $(l_i, a, b, r_j)$ and $9$ of the form $(l_i, c, r_j)$. We see both considerations at work here:
	
	\begin{enumerate}
		\item The paths of this form that $a$ lies on contain $4$ nodes, whereas those that $c$ lies on contain $3$ nodes, and
		\item All $9$ of the paths of this form that $a$ lies on overlap with those that $b$ lies on. This means that if $b$ also has a mole, then clearing $a$ will leave all of these paths compromised. On the other hand, if $c$ has a mole, then clearing it \emph{guarantees} that at least $6$ paths of this form are not compromised.
	\end{enumerate}

	In order to capture the complex interactions between nodes and the synergies (both positive and negative) that nodes exhibit when considered in groups, the semivalue approach needs to be considered in order to analyse the expected number of shortest paths ``sterilised'' after clearing a node. The semivalue for group stress centrality of a node $v$ with the $\beta$ function such that $\beta(k) = 1$ for $k = 1$ and $\beta(k) = 0$ otherwise gives us exactly the number of shortest paths that we will cure upon removing a mole at a given location, given that all other nodes are equally likely to be compromised. If $c$ is compromised, then removing a mole placed there will cure an expected number of $24 \frac{3}{4}$ shortest paths, whereas removing a mole at $a$ or $b$ will cure only an expected number of $16 \frac{3}{4}$ shortest paths.

\section{Game-Theoretic Node Centrality Measures Based on Connectivity}\label{section:connectivity}

\noindent In this section, we review the game-theoretic centralities that are based on connectivity. Typically, this means that the centralities are based on, or inspired by, the Myerson value and Myerson's \emph{graph-restricted game} \cite{Myerson:1977}. In this approach, the model consists of a graph and a value function defined over all subsets of nodes (who are treated as players in a cooperative game).

The next step is to construct a new cooperative game based on the above value function and graph. In this context, we will say that the network topology ``restricts'' the game, i.e., it is ``graph restricted''. For now, we will leave this as an abstract concept, and discuss its variations when we get to the specific centrality measures. 
In particular, in Section~\ref{subsection:graph:restricted:games}, we formally introduce the Myerson value and his graph-restricted game. In  Section~\ref{subsection:gomez:et:al}, we analyze the centrality proposed by \cite{Gomez:et:al:2003}, which is based on the Myerson value. Section~\ref{subsection:del:pozo:et:al} presents the centrality due to \cite{del:Pozo:2011} for digraphs. The centralities due to \cite{Amer:Gimenez:2007} and \cite{Belau:2014} then follow in Sections~\ref{amer:et:al:2007} and~\ref{subsection:belau:et:al:2014}, respectively. Section~\ref{subsection:grofman:owen:1982} discusses the Banzhaf network centrality, proposed by~\cite{Grofman:Owen:1982}, which shares certain (but not all) traits with the connectivity approach.  Finally, in Sections \ref{subsection:amer:connectivity} and \ref{subsection:kim:tackseung} we discuss alternative concepts of connectivity and graph-restricted games.

\subsection{The Myerson Value and its Computational Properties}\label{subsection:graph:restricted:games}
\noindent Graph-restricted games were introduced in Myerson's seminal work from \citeyear{Myerson:1977} \cite{Myerson:1977}.
In the model analysed by Myerson, a network consists of players (nodes of the graph) who can communicate either directly, via the links that directly connect them, or through intermediaries, i.e., when there exists a path in the network between them. This network is represented by a graph, $G$. A natural assumption, then, is that only players who communicate are able to cooperate. This idea was embodied in a coalitional game, by introducing a value function, $\nu: 2^V \rightarrow \mathbb{R}$, where $\nu(\emptyset) = 0$, that assigns to all coalitions a real value. However, disconnected coalitions cannot communicate and therefore achieve this value. In this context, a \textit{connected coalition} is defined as a coalition for which there exists a path between any two nodes belonging to the coalition, that visits nodes only within the coalition. In other words, a connected coalition is one that induces a connected subgraph. We will write $\mathcal{C}(G)$ to denote the set of all connected coalitions that can be made of nodes within $G$.

\cite{Myerson:1977} proposed that the value of a disconnected coalition $C$,  should be the sum of the maximal connected components of the subgraph induced by $C$. We will denote the set of all such connected components by $\mathcal{K}(C)$. Given this we can formulate the definition of the characteristic function for Myerson's graph-restricted game, which we denote by $\nu^{\mM}_G(S)$. Formally:
\begin{equation}
\label{equation:nu_G}
\nu^{\mM}_G(S) =
\begin{cases}
\nu(S) & \textnormal{ if } S \in \mathcal{C}(G) \\
\sum_{K_i \in \mathcal{K}(S)}\ \nu(K_i) & \textnormal{ otherwise.}  \\
\end{cases}
\end{equation}

Myerson's celebrated result \cite{Myerson:1977} is that the Shapley value of a node $v$ in this graph-restricted game, $\mathit{SV}_v(G;\nu^{\mM}_G)$, is equal to his solution concept---now known as the Myerson value---of $v$ in the original game, denoted by $\mathit{MV}_v(G;\nu)$. The Myerson value of node $v$ is formally and uniquely defined by the following axioms:



\begin{enumerate}
	\item \textbf{Efficiency in connected components:} The entire payoff of a connected component is distributed among the coalition members.
	
	Formally: $\forall_{K_i \in \mathcal{K}(V)} \sum_{v \in K_i} MV_v(G;\nu) = \nu(K_i)$;
	
	\item \textbf{Fairness:} Any two players that are neighbours are rewarded equally for the edge between them. In other words,  if we add an edge $(s,t)$ to $G$ (denoted by $G \cup (s,t)$), then the Myerson value of $s$ and $t$ changes by the same amount.
	
	Formally: $MV_s(G \cup (s,t); \nu) - MV_s(G; \nu) = MV_t(G \cup (s,t); \nu) - MV_t(G; \nu)$;
	
	\item \textbf{Stability:} if $\nu$ is superadditive, then adding an edge $(s,t)$ will not decrease the payoff of $s$ or $t$.
\end{enumerate}

In the remainder of this section, we discuss an algorithm to compute the Myerson value and then discuss how it can be used as a centrality measure.

\subsubsection{An Algorithm for Computing the Myerson Value}

A general algorithm for computing the Myerson value was proposed by \cite{Skibski:et:al:2014}. This algorithm runs in $O(|\mathcal{C}(G)||E|)$ time, where $|\mathcal{C}(G)|$ is the number of connected subgraphs of $G$. In general, this is difficult to improve upon, since the number of steps to consider the values of all subgraphs (which are given by the characteristic function $\nu$) is $O(|\mathcal{C}(G)|)$, and for every subgraph $O(|E|)$ steps are required to identify its set of neighbors. Interestingly, given a polynomial number of connected subgraphs, this algorithm also runs in polynomial time.

\RestyleAlgo{ruled} 
\begin{algorithm}[h!]
	\SetAlgoVlined
	\LinesNumbered

	\KwIn{Graph $G\hspace{-0.1cm}=\hspace{-0.1cm}(V,E)$, function $\nu: \mathcal{C}(G) \rightarrow \mathbb{R}$}
	\KwOut{Myerson value for game $\nu_G$}
	\vspace{0.05cm}
	$DFSMyersonV$ \Begin{
		\textit{sort nodes and list of neighbors by degree desc.;} \\
		\ForLine{$i \gets 1$ \KwTo $|V|$}{$MV_i(\nu_f) \gets 0;$} \\
		\For{$i \gets 1$ \KwTo $|V|$}{
			$DFSMyersonVRec(G, (v_i), \{v_i\}, \{v_1, \ldots, v_{i-1}\}, \emptyset, 1)$; \\
		}
	}
	\vspace{0.05cm}
	$DFSMyersonVRec(G, path, S, X, XN, startIt)$ \Begin{
		$v \gets path.last()$; \\
		
		\For{$it \gets startIt$ \KwTo $|E(v)|$}{
			$u \leftarrow \mN(v).get(it)$; \AlgorithmComment{$it$'s neighbor of $v$}\\
			\If{$u \not \in S \land u \not \in X$}{
				$DFSMyersonVRec(G, (path, u), S \cup \{u\}, X, XN, 1)$; \\
				$X \gets X \cup \{u\}$; $XN \gets XN \cup \{u\}$;
			} \ElseLine{\IfLine{$u \in X$}{$XN \gets XN \cup \{u\}$;}}
		}
		$path.removeLast()$; \\
		
		\If{$path.length() > 0$}{
			$startIt \gets \mN(path.last()).find(v)+1)$; \\
			$DFSMyersonVRec(G, path, S, X, XN,startIt)$;
		}
		\Else{
			\ForEach{$v_i \in S$}{
				$MV_i(\nu_G) \gets MV_i(\nu_G) + \frac{(|S|-1)!(|XN|)!}{(|S|+|XN|)!}\  \nu(S)$;\label{line:coefficient1}
			}
			\ForEach{$v_i \in XN$}{
				$MV_i(\nu_G) \gets MV_i(\nu_G) - \frac{(|S|)!(|XN|-1)!}{(|S|+|XN|)!} \ \nu(S)$;\label{line:coefficient2}
			}
		}
	}
	\caption{The DFS-based algorithm for calculating the Myerson value due to \cite{Skibski:et:al:2014}.}
	\label{algorithm:myersonvalue}
\end{algorithm}

The pseudocode for the algorithm due to \cite{Skibski:et:al:2014} is presented in Algorithm~\ref{algorithm:myersonvalue}. We briefly explain how it works. The basic idea is to traverse all connected coalitions using depth-first search (DFS). During this process, an internal variable---$S$---represents the ``current'' connected coalition. The main function---$DFSMyersonVRec$---is called recursively: once, where the node currently being processed by the DFS algorithm is included in $S$, and once, where that node is forbidden from ever being included in $S$. By recursively calling this function in such a manner, all connected coalitions are eventually traversed. During traversal for some connected coalition, $S$, the algorithm stores in $XN$ the neighbor set of $S$. The Myerson value of node $v$ is computed using Equation~\eqref{SV1}---the formula of the Shapley value---with the characteristic function for the graph-restricted game. Recall that Equation~\eqref{SV1} iterates over all permutations in $\Pi(N)$, and for every such permutation, $\pi$, computes the marginal contribution that the node $v$ makes to $C_\pi(v)$---the coalition of nodes preceding $v$ in the permutation $\pi$. The important observation here, is that when we take the graph-restricted characteristic function defined in Equation~\eqref{equation:nu_G}, the marginal contribution of $v$ to $C_\pi(v)$ becomes:
\begin{equation}\label{eqn:MC_Myerson}
\nu^{\mM}_G(C_\pi(v)\cup\{v\}) - \nu^{\mM}_G(C_\pi(i))  \ =\  \nu(K_1\cup\dots\cup K_d\cup\{i\}) - \nu(K_1) - \dots - \nu(K_d),
\end{equation}
where $K_1, \dots, K_d$ are the maximal connected subsets of $C_\pi(i)$ that are connected to $v$. In other words, if $v$ joins $C_\pi(v)$, then it connects these components, but not others. It is therefore possible to express the Myerson value as a weighted sum over connected subsets $S$ of the value $\nu(S)$. In order to do so, we must compute for each such connected subset how many times it appears in the sum in Equation \eqref{SV1}. Based on these observations, for every connected coalition, $S$, when computing the Myerson value of the node $v$, the algorithm distinguishes between the following two possibilities:

\begin{itemize}
	\item If $v\in S$, then $\nu(S)$ appears in the following part of Equation~\eqref{eqn:MC_Myerson}: $\nu(K_1\cup\dots\cup K_d\cup\{v\})$. This happens whenever $S=K_1\cup\dots\cup K_d\cup\{v\}$. More specifically, it happens in a permutation $\pi$ if and only if $S\setminus \{v\} \subseteq C_\pi(v)$ and $C_\pi(v) \cap XN = \emptyset$. Let us compute how many such permutations there are by constructing them. First of all, since the players in $N \setminus (XN \cup S)$ are completely irrelevant here, they can be placed in $\pi$ in any position. The number of ways in which this can be done is:
	\[
	\binom{|N|}{|N| - |XN \cup S|} (|N| - |XN \cup S|)! = \frac{|N|!}{|XN \cup S|!}.
	\]
	After placing those players, we still have $|S\cup NX|$ unassigned slots in $\pi$, which should be assigned as follows. The members of $S\setminus \{v\}$ should be placed in the first $(|S|-1)$ unassigned slots; there are $(|S|-1)!$ different ways in which this can be done. After that, the first of the remaining slots should be occupied by $v$, while the rest should be occupied by the members of $XN$; there are $|XN|!$ different ways to do this. The total number of permutations becomes: 
	\[
	\frac{|N|!(|S|-1)!|XN|!}{|XN \cup S|!}.
	\]
	Now, since Equation~\eqref{SV1} averages each marginal contribution over all $|N|!$ possible permutations, the value of $S\setminus \{v\}$ must be divided by this coefficient when computing the Myerson value of $v$. In other words, for any connected coalition $S$ that satisfies the above requirements, the value $\nu(S)$ appears the following number of times when computing the Myerson value:
	\[
	\frac{(|S|-1)!|XN|!}{|S \cup XN|!}.
	\] 
	\item On the other hand, if $v \notin S$, then $\nu(S)$ can only appear in the following part of Equation~\eqref{eqn:MC_Myerson}: $- \nu(K_1) - \dots - \nu(K_d)$. This happens whenever $S=K_j$ for some $j\in\{1,\dots,d\}$. More specifically, it happens in a permutation $\pi$ if and only if $S \subseteq C_\pi(v)$ and $C_\pi(v) \cap XN = \emptyset$ and $v\in XN$. Just as in the previous case, it is possible to compute the number of ways in which such a permutation can be constructed. We arrive at the following coefficient, which is the number of times $\nu(S)$ appears when computing the Myerson value of $v$:
	\[
	-\ \frac{|S|!(|XN| - 1)!}{|S \cup XN|!}.
	\]
\end{itemize}

The above two coefficients are used in lines \ref{line:coefficient1} and \ref{line:coefficient2} of Algorithm~\ref{algorithm:myersonvalue}. We provide a closed-form formula for the Myerson value based on these observations:

\begin{align}
MV_v(\nu) = &\sum_{v \in S \in \mathcal{C}(V)} \frac{(|S|-1)!|E(S)|!}{|S \cup E(S)|!} \nu(S \cup \{v\})\nonumber\\ - &\sum_{S \in \mathcal{C}(V), v \in E(S)} \frac{|S|!(|E(S)| - 1)!}{|S \cup E(S)|!} \nu(S),
\end{align}

where $E(S)$ denotes the neighbor set of $S$, i.e., $|E(S)| = XN$ from Algorithm~\ref{algorithm:myersonvalue}. We note that another expression that allows for polynomial computation of the Myerson value given a polynomial number of connected coalitions was developed by \cite{elkind:2014}. \citeauthor{elkind:2014} further characterised the class of graphs for which the number of connected coalitions is polynomial in $|V|$, thereby admitting polynomial computation of the Myerson value and other solution concepts.

This concludes the discussion of the computation of the Myerson value.

\subsubsection{The Myerson Value as a Centrality Measure}

\begin{figure}[thbp]
	\center
	\includegraphics[width=7cm]{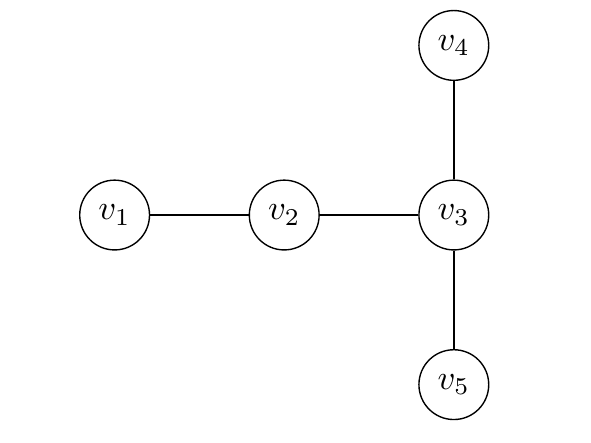}
	\caption{A sample network used to illustrate connectivity and how it impacts the Myerson value}
	\label{figure:Myerson}
\end{figure}

Since the Myerson value assigns values to individual nodes by taking into account the topology of the network, it can be thought of as a centrality measure, as demonstrated in the following example:

\begin{example}
	Let us consider the network presented in Figure~\ref{figure:Myerson} and the characteristic function $\nu(C) = |C|^2$. This particular characteristic function was chosen since it is superadditive, and it gains more and more value the more neighbours join. In particular, if no network were present, each player would have an equal Shapley value, which is $5$. However, as soon as we take the network topology into consideration, we find that not all players can participate as often in the larger (more rewarding) coalitions. Since the game is quadratic (making it strictly convex), rather than linear, this makes being in a \textit{central} position in the network even more beneficial. The Myerson values for the nodes $v_1, v_2, v_3, v_4$ and $v_5$ are $3.167, 6, 8.167, 3.333$ and $3.333$, respectively. Clearly, node $v_3$ receive the highest payoff, which reflects the fact that it plays the most important role in connecting the other nodes. For instance, adding $v_3$ to the coalition $\{v_2, v_4, v_5\}$ connects them, changing the value of the coalition from $3$ to $16$. 
\end{example}

Having introduced graph-restricted games due to Myerson, over the next five sections, we will now show how these games were used in the literature to construct different game-theoretic network centralities. 

\subsection{Centrality due to \cite{Gomez:et:al:2003}}\label{subsection:gomez:et:al}



\noindent\citefull{Gomez:et:al:2003} propose to use the difference between the Myerson value and the Shapley value of nodes within a network as a measure of centrality. Formally, this measure is given by:
\begin{equation}
\label{Gomez}
MV_{v}(G;\nu) - SV_{v}(\nu),
\end{equation}
where $MV_{v}(G;\nu)$ is the Myerson value of node $v$ and $SV_{v}(\nu)$ is its Shapley value. Intuitively, this measure takes the Shapley value, which accounts for the importance of nodes without considering the network topology, and subtracts it from the Myerson value, which accounts for the importance of nodes in a network context. Hence, this measure of centrality quantifies how much power a node has gained (or lost) due to its position in the network. However, note that when $\nu$ is symmetric, then $SV_{v}(\nu)$ is equal for all $v \in V$. In this case, the centrality measure is equivalent to the Myerson value (shifted by some constant). 

%
%

Since the Shapley value is simply used as an index of power of a player in the game, the authors note that the Banzhaf index could be used instead. Indeed, we note that any semivalue could be used depending on the application at hand. As such, an interesting research direction would be to study what these different power indices (i.e., solution concepts such as semivalues) would imply for the centrality measure; clearly, many properties of the measure rely not only on $\nu$ but also on the adopted solution concept.

%
%
%
%

\begin{example}
	To demonstrate the basic idea behind the centrality measure due to \cite{Gomez:et:al:2003}, consider again the network in Figure~\ref{figure:Myerson}. Here, given the characteristic function $\nu(C) = |C|^2$, the centrality of $v_1, v_2, v_3, v_4$ and $v_5$ is: $-2.167, 1, 3.167, -2.333$ and $-2.33$, respectively. Since this game is symmetric, these values are basically shifted Myerson values. It is possible, however, to define a game that is not symmetric (and in this sense includes information that is extraneous to the network), in which case the Gomez centrality can produce a ranking different to that produced by the Myerson value.
\end{example}

\subsection{Centrality in Directed Social Networks by~\cite{del:Pozo:2011}}\label{subsection:del:pozo:et:al}

\noindent\citefull{del:Pozo:2011} build upon their previous work \cite{Gomez:et:al:2003} presented in the above section and define a family of centrality measures for directed social networks. In the previous section,  \cite{Gomez:et:al:2003} used the Shapley value as a measure of power that disregards the topology of the network, whereas the Myerson value was used as a measure of power within the network. The difference between the two values was then used as a measure of how much power a node gains (or loses) due to its position in the network. \cite{del:Pozo:2011} extend this measure to digraphs by proposing their version of \emph{digraph-restricted games} (as opposed to graph-restricted games). Here, the Shapley value is used just as before, as a standard measure of power disregarding the network. However, the Myerson value is replaced with a generalised solution concept (and the digraph-restricted game is a generalised cooperative game), where the order of the players matters in determining its value.

This approach makes sense for digraphs, since an ordered group of nodes may communicate in a digraph only if each consequent node is connected. To achieve this, the authors take an underlying model of a digraph, $D$, and cooperative game $\nu$. They then produce a generalised cooperative game that they refer to as the digraph-restricted game, $\nu^{\mathcal{P}*}_D$, where ${\mathcal{P}}$ stands for Pozo, and develop a family of generalised solution concepts in order to account for the importance of nodes in the digraph-restricted game. This family of measures, referred to by $\chi_\alpha$, is parametrised by $\alpha$, which takes any value in $[0,1]$. On one extreme, when $\alpha = 0$, $\chi_\alpha$ is equal to a solution concept proposed by \cite{Nowak:Radzik:1994}, whereby players may only join an ordered coalition at the very end. On the other extreme, when $\alpha = 1$, $\chi_\alpha$, it is equal to a solution concept proposed by \cite{Sanchez:Bergantinos:1997}, whereby players can join an ordered coalition at any position (including in between other nodes). This consequently impacts the types of marginal contributions that players can make in a generalised cooperative game. Setting $\alpha$ anywhere between the two extremes results in a hybrid of the two aforementioned solution concepts. When $\alpha$ decreases, more value is given to the players that join an ordered coalition last. Let us now formally introduce the two generalised solution concepts below.

\paragraph{Solution Concepts for Generalised Cooperative Games:} Here, we briefly introduce two prominent solution concepts for generalised cooperative games.\footnote{The specific formulations we use here are different from the ones that were originally proposed by the authors of the solution concepts. The formulations used here can be found, for example, in the work by \cite{del:Pozo:2011}.}
To this end, let $\pi_C = \{c_1, c_2, \ldots, c_{|C|}\}$ be a permutation of $C$. We will denote by $\pi(i \rightarrow l)$ a permutation of $C \cup \{i\}$ such that $i$ is inserted in $\pi$ immediately after the $l$'th player (if $l = 0$, then $i$ is inserted as the first player). Formally, $\pi(i \rightarrow l) = \{c_1, c_2, \ldots, c_l, i, \ldots , c_{|C|}\}$. Let us first introduce the solution concept, proposed by \cite{Sanchez:Bergantinos:1997}. This solution concept, now known as the Sanchez-Bergantinos value, is computed for a player $i$ in a game with a set of $n$ players, $I$, and defined by the generalised characteristic function $\nu^*$, as follows:


\begin{align}\label{equation:sanchez:bergantinos}
\Psi_i^{SB}(\nu^*) &= \sum_{\pi \in \Pi(C), C \subseteq I \setminus \{i\} } \frac{(n - |C| - 1)!}{n!(|C| + 1)} \sum_{l = 0}^{|C|} (\nu^*(\pi(i \rightarrow l)) - \nu^*(\pi))\nonumber \\
&=  \sum_{\pi_S \in \Omega(n), i \in S} \frac{\Delta^*_{\nu^*}(\pi_S)}{|S|!|S|},
\end{align}

where $\Delta^*_{\nu^*}(\pi_S)$ denotes the generalised dividends, discussed earlier in Section~\ref{ch:3-preliminaries} and defined formally in Equation~\eqref{eqn:deltaStar}
 and $\Omega(I)$ is the set of all ordered coalitions of $I$, i.e., $\Omega(I) = \bigcup_{C \subseteq I} \Pi(C)$. The other solution concept we wanted to discuss is the Nowak-Radzik value, which was proposed by \cite{Nowak:Radzik:1994}. This solution concept is computed for a player $i$ in a game with a set of $n$ players, $I$, and defined by the generalised characteristic function $\nu^*$, as follows:

\begin{align}\label{equation:nowak:radzik}
\Psi_i^{NR}(\nu^*) &= \sum_{\pi \in \Pi(C), C \subseteq I \setminus \{i\} } \frac{(n - |C| - 1)!}{n!} (\nu^*(\pi(i \rightarrow |C|)) - \nu^*(\pi)) \nonumber\\
&= \sum_{\pi_S \in \Omega(I), \pi_S(i) = |S|} \frac{\Delta^*_{\nu^*}(\pi_S)}{|S|!}.
\end{align}

The difference between these two solution concepts is that the former distributes the generalised dividend among all players in $\pi_S$ equally (and hence the requirement $i \in S$), whereas the latter gives the entire generalised dividend to the last contributing player in $\pi_S$ (and hence the requirement $\pi(i) = |S|$, i.e., that $i$ is the last player). Having defined the two solution concepts, we will now formally define the centrality measure due to \cite{del:Pozo:2011}, which builds upon a parametrised hybrid of the two. 

\paragraph{Centrality According to \cite{del:Pozo:2011}}: The authors define a family of solution concepts for generalised cooperative games, parametrised by $\alpha$, as follows:

\begin{equation}
\Psi_i^\alpha(\nu^*) = \sum_{\pi_S \in \Omega(I), i \in S} \Delta^*_{\nu^*}(\pi_S) \frac{\alpha^{|S| - \pi_S(i)}}{|S|!\sum_{j = 0}^{|S| - 1}\alpha^j}.
\end{equation}

Note that we abuse notation and assume that $0^0 = 1$. From the above equation, it is clear that this family of solution concepts is in fact a hybrid of Equations \eqref{equation:sanchez:bergantinos} and \eqref{equation:nowak:radzik}. Specifically, when $\alpha = 0$ we obtain the \cite{Nowak:Radzik:1994} value as in Equation~\eqref{equation:nowak:radzik}, and when $\alpha = 1$ we obtain the \cite{Sanchez:Bergantinos:1997} value, as in Equation~\eqref{equation:sanchez:bergantinos}. Between the two extremes, a smaller $\alpha$ implies that a higher proportion of a dividend is gradually given to the later players in an ordered coalition, whereas a larger $\alpha$ implies a more even distribution of power regardless of the order of players.

We are almost ready to introduce the measure proposed by \cite{del:Pozo:2011}. The last definition we need to introduce is that of \emph{digraph-restricted games}. To this end, let $\nu^{\mathcal{P}*}_D$ denote the game $\nu$ restricted by the digraph $D$, and let $\text{Paths}(D)$ denote the set of all paths without loops in the digraph $D$. We will first show how games belonging to the unanimity basis can be restricted by digraphs (see Section~\ref{ch:3-preliminaries} for more on unanimity bases). 
Recall that when $u^*_S$ is a generalised characteristic function for a (regular, not generalised) unanimity game for the coalition $S$, then we have $u^*_S = \sum_{\pi_S \in \Pi(S)} w^*_{\pi_S}$, where $w^*_{\pi_S}$ is the generalised unanimity game for $\pi_S$. The digraph restriction of the unanimity game $u^*_S$ according to \citeauthor{del:Pozo:2011}, denoted by $u^{\mathcal{P}*}_{D,S}$, is defined as follows:
\[
u^{\mathcal{P}*}_{D,S} = \sum_{\pi_S \in \Pi(S) \cap \text{Paths}(D)} w_{\pi_S}.
\]
In other words, only consecutively connected ordered coalitions (i.e., those that are paths) that contain every member of $S$ have value $1$.
The general formula for digraph restricted games not belonging to the unanimity basis, then, is as follows:
\[
\nu^{\mathcal{P}*}_D = \sum_{S \in V} \Delta_{\nu}(S) u^{\mathcal{P}*}_{D,S},
\]
where $\Delta_{\nu}(S)$ are the regular (not generalised) Harsanyi dividends for the cooperative game $\nu$. Note that---unlike $\nu$---$\nu^{\mathcal{P}*}_D$ is a generalised characteristic function. For this reason, solution concepts for generalised cooperative games must be used as a measure of power. Now, we are ready to present the centrality measure due to \cite{del:Pozo:2011}. Given a digraph $D$ and a (regular, not generalised) cooperative game $\nu$, the centrality of a node $v$ is: 
\begin{equation}
\Psi_v^\alpha(\nu^{\mathcal{P}*}_D) - SV_v(\nu).
\end{equation}

\begin{figure}[th]
	\center
	\includegraphics[width=7cm]{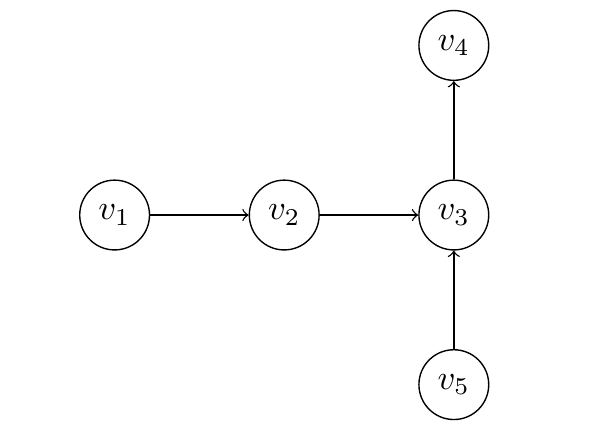}
	\caption{A sample digraph used to compare a number of centrality measures based on connectivity.}
	\label{figure:del:Pozo}
\end{figure}

\begin{example}
	Consider the directed network in Figure~\ref{figure:del:Pozo} along with the characteristic function $\nu(C) = |C|^2$. Let us start by calculating the Harsanyi dividends for this game. Since $\nu$ is symmetric, we will write $\Delta(|C|)$ to be equivalent to $\Delta_\nu(C)$ (which will allow us to define the dividends for $5$ possible coalition sizes, rather than for all $2^5$ coalitions). First, we have $\Delta(1) = 1^2 = 1$. Next, $\Delta(2) = 2^2 - 2 \Delta(1) = 2$. Moving on, $\Delta(3) = 3^2 - 3 \Delta(1) - 3 \Delta(2) = 0$, $\Delta(4) = 4^2 - 4 \Delta(1) - 6 \Delta(2) - 4 \Delta(3) = 0$, and finally $\Delta(5) = 5^2 - 5 \Delta(1) - 10 \Delta(2) - 10 \Delta(3) - 5 \Delta(4) = 0$. The Shapley value of any player is simply $\Delta(1) + \frac{4 \Delta(2)}{2} = 1 + \frac{4}{2} = 5$, since every node belongs to only one coalition of size $1$ and $4$ coalitions of size $2$. Alternatively, the value of the grand coalition is $25$, and since the game is symmetric, then this value is split equally between all players, i.e., $\frac{25}{5} = 5$ each.
	
	We will now express $\nu^{\mathcal{P}*}_D$ given the digraph $D$ from Figure~\ref{figure:del:Pozo}. We have $\nu^{\mathcal{P}*}_{D}(\pi_C) = \sum_{S \subseteq V} \Delta_{\nu}(S) u^{\mathcal{P}*}_{D,S}(\pi_C)$. Recall that $u^{\mathcal{P}*}_{D,S}(\pi_C) = 1$ if and only if $\pi_C$ is a path in $D$ and $C$ contains all the players from $S$. Since the Harsanyi dividends are equal to $0$ for all coalitions of size greater than $2$, our job is not too difficult. $\nu^{\mathcal{P}*}_{D}(\pi_C) = |C| + 2 \times [\text{number of paths in } D \text{ of length } 2 \text{ in } \pi_C]$. In terms of the unanimity basis for generalised cooperative games, we have: $\nu^{\mathcal{P}*}_D = w^*_{\{1\}} + w^*_{\{2\}} + w^*_{\{3\}} + w^*_{\{4\}} + w^*_{\{5\}} + 2 \times [w^*_{\{1,2\}} + w^*_{\{2,3\}} + w^*_{\{3,4\}} + w^*_{\{5,3\}}]$. This form of $\nu^{\mathcal{P}*}_D$ is very convenient, as we now have the coefficients in the unanimity basis, and hence the dividends required to calculate the new centrality.
	
	The Nowak-Radzik value vector for the nodes $v_1, v_2, v_3, v_4, v_5$ is therefore $1, 2, 3, 2, 1$, whereas Sanchez-Bergantios value vector is $1.5, 2, 2.5, 1.5, 1.5$, and the centrality due to \cite{del:Pozo:2011} is simply a shift of these values by $-5$. Note that the sum of the centralities is equal for both solution concepts; only the distribution has changed. We can now see how important the edges and their direction is for the Nowak-Radzik value. For example, the benefit of the edge $v_3 \rightarrow v_4$ is split equally between nodes $v_3$ and $v_4$ in the Sanchez-Bergantios value, however in the Nowak-Radzik value $v_3$ gains nothing from it, and the entire worth of the edge is given to the last node---$v_4$.
\end{example}

The complexity of this measure has not been extensively studied. The characterisation of the centrality measure given by \cite{del:Pozo:2011} requires the calculation of all Harsanyi dividends for all paths in the graph. A naive algorithm would have to traverse all paths without loops in a graph, possibly resulting in $O(\sum_{k = 1}^n \binom{n}{k} k!)$ calculations. Furthermore, calculating all Harsanyi dividends requires $O(\sum_{k = 1}^n \binom{n}{k} 2^k)$ calculations. It is an open problem whether these complexities can be improved upon in general or for certain classes of graphs.


One criticism of the above family of centrality measures is as follows: why does it not allow---through the parameter $\alpha$ or otherwise---for the option of the first nodes in a path to benefit most? It is purely by convention that we equate paths with ordered coalitions such that the source of the path is the first node in the coalition, and the destination is the last. This can, of course, be remedied by reversing the edges in the graph depending on the application, but it is not an inherent parameter of the centrality. We will return to this question in Section \ref{subsection:VL}; the centrality measure presented therein does the opposite of the Nowak-Radzik value, giving more value to nodes at the beginning of paths.

In the next section, we move on to a measure of centrality developed by \cite{Amer:Gimenez:2007}. Although the centrality presented in this section stems from the work by \cite{Gomez:et:al:2003}, which makes prominent use of the Myerson value, it is not in fact based on the idea of graph-restricted games according to Myerson. The idea of digraph-restricted games presented here is somewhat convoluted, and the result is not intuitive. The centrality in the next section, however, is based on almost a direct translation of Myerson's graph-restricted games to digraphs.

\subsection{Accessibility in Digraphs and Centrality due to \cite{Amer:Gimenez:2007}}\label{amer:et:al:2007}
%

\noindent\citefull{Amer:Gimenez:2007} define the notion of accessibility in digraphs based on an idea similar to Myerson's graph-restricted games. Although they do not use the term ``centrality'', they do talk about the ranking of nodes in a network. In essence, their notion of accessibility is a centrality measure suitable for applications in which it is important for central nodes to be reachable (i.e., accessible). In a sense, this is the opposite of closeness centrality, where it is desirable to be able to reach other nodes quickly \emph{from} central nodes. \cite{Amer:Gimenez:2007} define a digraph-restricted, generalised cooperative game on a directed network, in much the same way that Myerson defined a graph-restricted cooperative game on networks.
The approach taken is to partition ordered coalitions into a maximal set of smaller ordered coalitions that form paths in the digraph, and to equate their value with the sum of the values of these maximal coalitions.
Formally, let $\Pi(V)$ be the set of all permutations on a set of nodes, $V$. Furthermore, let $\pi \in \Pi(V)$ be a permutation of $V$ and denote by $\pi(j)$ the position of node $j$ in $\pi$, and by $\pi_j$ the $j$'th element in $\pi$. Next, we must introduce connected consecutive subcoalitions of $\pi$ according to a digraph $D$. We say $\pi = \{\pi_p, \pi_{p+1}, \ldots, \pi_{p+u}\}$ is such a coalition whenever $(\pi_j, \pi_{j+1}) \in E(D)$. Additionally, if $p = 1$ or $(\pi_{p-1}, \pi(p)) \not\in E(D)$, then $\pi$ cannot be expanded to the left. If $p + u = |V|$ or $(\pi(p+u), \pi(p+u+1)) \not\in E(D)$, then $\pi$ cannot be expanded to the right. If $\pi$ cannot be expanded to the left nor to the right, then it is a maximal connected consecutive subcoalition.
We denote by $\pi / D$ the set of maximal consecutive subcoalitions of $\pi$ according to $D$.

Let us now define the \textit{digraph-restricted generalised characteristic functio}n of a (regular, non-generalised) cooperative game that makes use of the above concepts. Recall that Equation~\eqref{SV1}---which defines the Shapley value---considers different permutations of players in each coalition. Now, according to \cite{Amer:Gimenez:2007}, in the case of digraphs, the characteristic function needs to be redefined, because some of the aforementioned permutations may not be feasible (i.e., not all of them form paths and may therefore not be able to communicate). 
We will denote by $\nu^{\mA\mG*}_D$ the generalised characteristic function obtained from restricting $\nu$ by the digraph $D$. For an ordered coalition, $\pi_S \in \Pi(S)$, we define:
\begin{equation}
\nu^{\mA\mG*}_D(\pi_S) = \sum_{\pi_C \in \pi_S / D} \nu(C).
\label{eq:digraph_modified}
\end{equation}
Note that $\nu^{\mA\mG*}_D$ is defined over \emph{ordered} coalitions, as opposed to $\nu$, which is defined over unordered coalitions.

The authors used these digraph-restricted games to characterise the \emph{accessibility} of a node in much the same way as Myerson can be equated to the Shapley value of the graph-restricted game. The accessibility of a node $v$ in the digraph $D$ with the cooperative game $\nu$, $AC_v(D;\nu)$ is defined as the Nowak-Radzik value (as defined in the previous section) of the digraph-restricted game:

\begin{equation}
AC_v(D; \nu) = \frac{1}{|V|!} \sum_{\pi \in \Pi(V)} [\nu^{\mA\mG*}_D(\pi_{|v} \cup v) - \nu^{\mA\mG*}_D(\pi_{|v})],
\label{eq:accessibility_shapley}
\end{equation}
where $\pi_{|v}$ is the ordered coalition $\pi$ cut off just before the node $v$ (i.e., if $\pi = \{v_1, v_2, \ldots, v_{k-1}, v_k, v, \ldots, v_n\}$, then $\pi_{|v} = \{v_1, v_2, \ldots, v_{k-1}, v_k \}$), and where $\pi \cup v$ is the ordered coalition with $v$ added at the end of $\pi$.

\begin{example}
	Let us consider the centrality due to  \cite{Amer:Gimenez:2007} with the characteristic function $\nu_2(C) = |C|^2$ and the digraph presented in Figure~\ref{figure:del:Pozo}---the same directed graph for which we calculated the centrality due to \cite{del:Pozo:2011} in the previous section. The accessibility of the nodes $v_1, v_2, v_3, v_4, v_5$ are $1, 1.4, 1.9, 1.63, 1$, respectively. One can see that the paths ending at $v_4$ are---on average---longer than those ending at $v_3$. However, more short paths exist that end at $v_3$ than those that end at $v_4$. To give more value to longer paths, one might consider the game $\nu_{10}(C) = |C|^{10}$. Now, the resulting ranks are $1$, $205.4$, $3259.9$, $21430.6$ and $1$, respectively. The exponent---in effect---specifies the extent of the synergy that happens when large groups collaborate. In this case, the longer paths that end with $v_4$ dominate the larger number of short paths ending with $v_3$. We saw the opposite for $\nu_2$, where synergy between nodes resulted in less value.
\end{example}

Accessibility satisfies the following properties:
\begin{itemize}
	\item Linearity: for any node, $v$, any pair of characteristic functions, $\mu$ and $\nu$, and any pair of real numbers, $x$ and $y$, we have: $AC_v(D; x\nu + y\mu) = x\cdot AC_v(D; \nu) + y\cdot AC_v(D; \mu)$;
	\item Dummy-player: \cite{Amer:Gimenez:2007} use a somewhat broader definition of the classic dummy player axiom. It is defined such that for any node $v$ that contributes exactly its singleton value to any coalition we have: $AC_v(D;\nu) = \nu^{\mA\mG*}_D(\{v\})$;
	\item Average efficiency: $\sum_{v} AC_v(D;\nu) = \frac{1}{|V|!} \sum_{\pi \in \Pi(V)} \nu^{\mA\mG*}_D(\pi)$;
	\item Given a complete digraph $D$ (i.e., one in which all possible edges are present), for every node $v\in V$, $AC_v(D; \mu)$ coincides with the Shapley value of $v$ in the game $(I,\nu)$;
	\item If a node $v$ is inaccessible in $D$ (i.e., if the indegree of $v$ equals $0$), then $AC_v(D;\nu) = \nu^{\mA\mG*}_D(\{v\})$;
	\item For any node $v$ in any digraph $D$, $AC_v(D; \nu)$ does not change by adding to $D$ an edge from $v$ to any other player;
	\item If $\nu$ is superadditive, then adding an edge from some node to the node $v$ does not decrease the value $AC_v(D; \mu)$.
\end{itemize}
These properties, however, do not uniquely characterise the accessibility measure. In contrast to the Shapley value, no set of axioms have yet been identified that would fully characterise accessibility.





The accessibility centrality measure was extended by \cite{Amer:Gimenez:2010} to something akin to that of semivalue as follows:
%
\begin{align}
&AC_v(D; \nu, \phi^\beta) = \phi^\beta_v(\nu_D) =\nonumber\\ &\sum_{C \in V \setminus \{v\}} \beta(|C|)\frac{(|V|-1-|C|)!(|C|)!}{|V|!} \sum_{\pi \in \Pi(C)} [\nu^{\mA\mG*}(\pi \cup \{v\}) - \nu^{\mA\mG*}(\pi)],
\label{eq:accessibility_semivalue}
\end{align}
where $\phi^\beta$ is the semivalue with which the probability of a coalition of size $k$ being chosen is equal to $\beta(k)$. Now, let us say that a game $(I,\nu)$ is \emph{zero-normalized} whenever $\nu(\{i\})=0$ for all $i\in I$. Then, the accessibility centrality measure defined in Equation~\eqref{eq:accessibility_semivalue} satisfies the following properties:

\begin{itemize}
	\item If $\nu$ is zero-normalized, then for every \emph{inaccessible} node $v$ (i.e., a node whose indegree is $0$), we have: $AC_v(D; \nu, \phi^\beta) = 0$;
	\item For every digraph $D=(V,E)$ and every node $v\in V$, if an edge $(v,u)$ is added to $E$, then the accessibility of $v$ stays the same, i.e., $AC_v((V, E); \nu, \phi^\beta) = AC_v((V, E \cup (v, u)); \nu, \phi^\beta)$;
	\item If $\nu$ is monotonic and zero-normalized, then for every digraph $D=(V,E)$ and every node $v\in V$, adding an edge $(u,v)$ will not decrease the accessibility of $v$. That is, $AC_v((V, E); \nu, \phi^\beta) \leq AC_v((V, E \cup (u, v)); \nu, \phi^\beta)$;
	\item Given a complete digraph $D$ (i.e., one in which all possible edges are present), for every node $v\in V$, $AC_v(D; \nu, \phi^\beta)$ coincides with $\phi^\beta_v(\nu)$---the semivalue of $v$ in the game $(V,\nu)$;
\end{itemize}

To date, the complexity of this measure has not been extensively studied, however a naive algorithm would have to traverse all ordered coalitions, of which there are $O(\sum_{k = 1}^n \binom{n}{k}k!)$. In the next section, we discuss a recent measure of centrality proposed by \cite{Belau:2014}. As opposed to the previous measures, this centrality is tailored for weighted networks. It makes use of the graph-restricted game by Myerson, but in a much different manner from the centrality measures presented up until now. This will be the last measure that we discuss that features Myerson's notion of connectivity.

\subsection{Cohesion Centrality due to~\cite{Belau:2014}}\label{subsection:belau:et:al:2014}

\noindent In this section we introduce cohesion centrality \cite{Belau:2014}. The defining feature of this node centrality measure is that it goes through the intermediary step of ranking all edges in terms of their importance for the \emph{cohesion} of the network. The underlying model for this measure consists of a weighted (or unweighted) digraph and a cooperative game on the set of nodes. From this, a new cooperative game \emph{on the set of edges} is defined, which is meant to represent how important each set of edges is for the cohesion, or interconnectedness, of the network. Next, either the Shapley value or the Banzhaf index is used to assess the importance for cohesion of every edge. The rank of each edge and its weight are then combined and---finally---one of Freeman's \cite{Freeman:1979} measures (degree, closeness or betweenness) is used to determine the weight of each node.

The first step of the cohesion centrality is to assign new weights to the edges in the network $D = (V,E)$. This is done according to the \emph{link-game} \cite{Meessen:1988, Borm:et:al:1992}---$\nu^{\mathcal{B}}_V: 2^E \rightarrow \mathbb{R}$---of a cooperative game $\nu: 2^V \rightarrow \mathbb{R}$, which is defined as follows:

\[
\nu^{\mathcal{B}}_V(S_E) := \nu^{\mM}_{(V, S_E)}(V),
\]

where $S_E \in E$ and $\nu^{\mM}_{(V, S_E)}$ is Myerson's graph-restricted game for $\nu$ and the graph $(V, S_E)$. In short, the link game defines a process through which we take a cooperative game ($\nu$) on the set of nodes ($V$), and from it define a cooperative game ($\nu^{\mathcal{B}}_V$) on the set of edges ($E$). Myerson's graph-restricted game is used to asses the value of the set of all nodes (i.e., the whole network) under different subsets of edges. The next step in defining the cohesion centrality of nodes is to use either the Shapley value or Banzhaf index to measure the value of each edge. The weight of the edges and their payoffs are then normalised to sum up to $1$ (this is done by dividing the payoff of each edge by $|E|$ and its weight by the sum of all weights) and combined according to the formula:

\[
\alpha \omega(s,t) + (1 - \alpha) \phi_{(s,t)}(\nu^{\mathcal{B}}_V),
\]

for some parameter $\alpha$ and where $\phi_{(s,t)}$ is either the Shapley value or Banzhaf index of the edge $(s,t)$ in the link game, and $\omega(s,t)$ is the weight of the edges $(s,t)$. Finally, one of the classical centralities defined by \cite{Freeman:1979} is used to define the cohesion centrality.

%
%
%
%
%

\begin{figure}[thbp]
	\center
	\includegraphics[width=7cm]{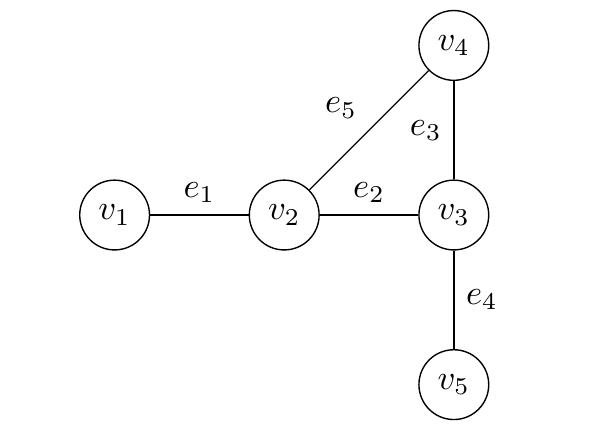}
	\caption{A sample weighted graph used to illustrate cohesion centrality due to \cite{Belau:2014}.
	}
	\label{figure:Belau}
\end{figure}

\begin{example}
	Let us consider the cohesion centrality of the nodes in Figure~\ref{figure:Belau}. Recall, that in order to define the underlying model we need both a graph and cooperative game on the set of nodes. For the latter, we will consider the cohesion game defined by \cite{Belau:2014}, which is defined as follows: $c(S,G) = \sum_{v \in S} c_v(G)$, where $c_v(G)$ is the number of nodes reachable from the node $v$ in the graph $G$. The link game (on the set of nodes) for this cohesion game then follows: $c^{\mathcal{B}}_V(S_E) = \sum_{v \in V} c_v((V, S_E))$, where $S_E \subseteq E$. The Shapley value of the edges $e_1$, $e_2$, $e_3$, $e_4$, and $e_5$ in the cohesion link game is $3.7$, $3.6$, $2.7$, $3.7$, and $2.7$, respectively. By using Freeman's Degree Centrality with these weights, we obtain a ranking of individual nodes. The rank of the nodes $v_1$, $v_2$, $v_3$, $v_4$, and $v_5$ is $3.7$, $10$, $10$, $5.4$, and $3.7$, respectively. We note that---if removed---$v_2$ and $v_3$ both disconnect one node from the rest of the network, making them the most important nodes for the cohesion of the network. Nodex $v_1$ and $v_5$ both only have one edge, which connects them to the rest of the network, making them the least important. Node $v_4$, on the other hand, has two edges, but they are not particularly important towards cohesion, making $v_4$ not much more important than $v_1$ or $v_5$.
\end{example}


In the next section, we discuss the centrality measure due to \cite{Grofman:Owen:1982}. This centrality considers all the paths within a directed network (just like the centralities due to \cite{del:Pozo:2011} and \cite{Amer:Gimenez:2007}), but it considers connectivity in a slightly different manner than the centrality measures that we have discussed up until now.

\subsection{Banzhaf Centrality due to \cite{Grofman:Owen:1982}}\label{subsection:grofman:owen:1982}

\noindent In this section we discuss the Banzhaf path-based centrality \cite{Grofman:Owen:1982}. This approach to connectivity is somewhat different than what we have discussed thus far, as it does not involve graph-restricted games of any manner. The centrality considers all connected ordered coalitions (i.e., paths without loops), much like the centrality due to \cite{del:Pozo:2011}, but without an explicit evaluation of groups of nodes. Rather, the authors of the measure go back to the roots of the Banzhaf index \cite{Banzhaf:1965}, which was defined as the relative number of possible swing votes that players have in weighted voting games.

\textbf{Banzhaf Index of Power for weighted voting games:} The Banzhaf power index is a very prominent solution concept in cooperative game theory. Originally proposed for voting games, it is defined as the probability that a player can change the outcome of an election. Here, we will give a slightly more general definition for weighted voting games (the original definition is equivalent to the one below, where all weights are equal to $1$).

In a weighted voting game, players attempt to pass a motion. A coalition is considered \emph{winning}, if the motion is carried. Consider a set $I$ of $n$ players, $\{1, \ldots, n\}$, where player $i$ has weight $w_{i}$, $0 < w_{i} \leq 1$ and $\sum_{i = 1}^n w_{i} = 1$. We define the quota, $q$, with $0 < q \leq 1$, as the sum of the weights of votes that is necessary for the motion to be carried. A coalition is \emph{winning} if the sum of the weights of its members is equal to or greater than $q$.

A $swing$ vote for player $i$ is a winning coalition that contains $i$ (i.e., $i$ was one of the players that voted for the motion to be carried) and his or her defection would change the outcome (the motion would not be carried if $i$ left it). Let the number of such coalitions for $i$ be denoted by $S_{i}$. The relative Banzhaf power index for player $i$, $B_{i}$, is then defined as
\begin{equation}
\label{B1}
B_i = \frac{S_{i}}{\sum_{i = 1}^{n} S_{i}}.
\end{equation}
Since such an index is game-specific (i.e., it heavily depends on the number of players), a total/normalised Banzhaf index, $B_{i}'$, is defined for comparison of nodes between games.

\begin{equation}
\label{B2}
B_i' = \frac{S_{i}}{2^{n} - 1}.
\end{equation}


The Banzhaf index is defined as the expected contribution of a player in a cooperative game.
 Consider the cooperative game $\nu$ for a weighted voting game such that $\nu(C) = 1$ if $C$ is winning and $\nu(C) = 0$ otherwise. Given this definition, the Banzhaf index of $\nu$ coincides with the Banzhaf index for the weighted voting game.

\textbf{Banzhaf centrality due to \cite{Grofman:Owen:1982}:} The idea behind this centrality measure is to count the number of times when a path is irreparably broken by the removal of a node. If the removal of $v$ from a path, $\pi$, means that no subset of the remaining nodes can form another path between the source and destination nodes, then $v$ clearly plays a significant role in $\pi$. The number of times a node can ``break'' a path in such a way determines its rank. 

%
More formally, let us defined the winning and losing coalitions. 
First and foremost, it is important to note that \cite{Grofman:Owen:1982} consider coalitions as ordered sets. The set of winning coalitions, then, is defined as the set of simple paths (paths without cycles) of length $2$ or greater. Thus $(1, 2, 3)$ represents a path from node $1$ to node $3$, whereas $(1, 3, 2)$ a path from node $1$ to $2$. A node $v$ has a swing vote in a winning coalition (i.e., path) $\pi$ that visits $v$ whenever the set of nodes that are visited by $\pi$ without $v$ can in no way form another path from the source to the destination. With these definitions of swing and winning coalition, the traditional definition of the Banzhaf power index (Equations \ref{B1} and \ref{B2}) can be applied directly.

As a critique of this approach, its connection to cooperative game theory and the Banzhaf index is slightly forced. Why do the authors go through modifying the original Banzhaf setting of weighted voting to networks? From a purely network science point of view, it seems perfectly reasonable to consider the number of paths that a node is essential to as its centrality measure and no new insight is brought from an analogy to the Banzhaf index.

\begin{example}
	Consider again the nodes in Figure~\ref{figure:Belau} and their Banzhaf path-based centrality \cite{Grofman:Owen:1982}. In particular, we will consider a variant of the centrality that was proposed by the authors, where swing votes of the first and last nodes in a path are not counted. After all, it is obvious that if the source or destination is removed, then the other nodes will not be able to reach these nodes; a source or destination node \emph{always} has a swing vote.
	
	Given the graph in Figure~\ref{figure:Belau}, the centrality of each of the nodes $v_1$, $v_4$ and $v_5$ according to \cite{Grofman:Owen:1982} is $0$. Indeed, nodes $v_1$ and $v_5$ are simply connected by one edge to the rest of the graph and lead nowhere else, making it clear that they do not have a single swing vote. As for $v_4$, it brings no indispensable edges to the graph. For all paths that it belongs to, it can be bypassed. In contrast, nodes $v_2$ and $v_3$ have a centrality of $\frac{12}{24}$ each. This is because each of them is indispensable for $12$ paths, and there are a total of $24$ swing votes. 
	More specifically, node $v_2$ is indispensable to the following paths: 
	\begin{align*}
	&v_1 \rightarrow v_2 \rightarrow v_3, &v_1 \rightarrow v_2 \rightarrow v_4 \rightarrow v_3,\\
	&v_1 \rightarrow v_2 \rightarrow v_3 \rightarrow v_4, &v_1 \rightarrow v_2 \rightarrow v_4,\\
	&v_1 \rightarrow v_2 \rightarrow v_3 \rightarrow v_5, &v_1 \rightarrow v_2 \rightarrow v_4 \rightarrow v_3 \rightarrow v_5,\\
	&v_3 \rightarrow v_2 \rightarrow v_1, &v_3 \rightarrow v_4 \rightarrow v_2 \rightarrow v_1,\\
	&v_4 \rightarrow v_3 \rightarrow v_2 \rightarrow v_1, &v_4 \rightarrow v_2 \rightarrow v_1,\\
	&v_5 \rightarrow v_3 \rightarrow v_2 \rightarrow v_1, &v_5 \rightarrow v_3 \rightarrow v_4 \rightarrow v_2 \rightarrow v_1.
	\end{align*} On the other hand, node $v_3$ is indispensable to the following paths:
	\begin{align*}
	&v_4 \rightarrow v_3 \rightarrow v_5, &v_4 \rightarrow v_2 \rightarrow v_3 \rightarrow v_5,\\
	&v_1 \rightarrow v_2 \rightarrow v_3 \rightarrow v_5, &v_1 \rightarrow v_2 \rightarrow v_4 \rightarrow v_3 \rightarrow v_5,\\
	&v_2 \rightarrow v_3 \rightarrow v_5, &v_2 \rightarrow v_4 \rightarrow v_3 \rightarrow v_5 \rightarrow,\\
	&v_5 \rightarrow v_3 \rightarrow v_4, &v_5 \rightarrow v_3 \rightarrow v_2 \rightarrow v_4,\\
	&v_5 \rightarrow v_3 \rightarrow v_2 \rightarrow v_1, &v_5 \rightarrow v_3 \rightarrow v_4 \rightarrow v_2 \rightarrow v_1,\\
	&v_5 \rightarrow v_3 \rightarrow v_2, &v_5 \rightarrow v_3 \rightarrow v_4 \rightarrow v_2.
	\end{align*}
\end{example}

The complexity of computing this centrality has not been studied in the literature. A naive algorithm considers all paths within the network (of which there are potentially $O(\sum_{k = 1}^n \binom{n}{k}k!)$), and for each node in the path checks whether the rest of the nodes can rearrange themselves in order to form a different path between the same pair of nodes.

In the following section, we will discuss two other variatoions of connectivity that have also been applied to game-theoretic network centrality.

\subsection{Connectivity Games due to \cite{Amer:Gimenez:2004}}\label{subsection:amer:connectivity}
\noindent In approaching connectivity, \cite{Amer:Gimenez:2004} formalised an approach that differs from Myerson's graph restricted games in its evaluation of disconnected coalitions. Whereas Myerson \cite{Myerson:1977} assumed that disconnected coalitions cannot fully cooperate and therefore their value should be the sum of their maximally connected parts, \cite{Amer:Gimenez:2004} postulate that the value of such coalitions should be $0$. In line with this, \cite{Amer:Gimenez:2004} defined the following characteristic function of a simple game:\footnote{\footnotesize Simple coalitional games are a popular class of games, where every coalition has a value of either 1 or 0.}
\begin{equation}
\label{equation:nu_f}
\nu^{\mA\mG}_G(S) =
\begin{cases}
1 & \textnormal{if } S \in \mathcal{C}(G) \\
0 & \textnormal{ otherwise, }  \\
\end{cases}
\end{equation}
\noindent where $\mA\mG$ stands for Amer and Gimenez.
The game is referred to as the \textit{$0$-$1$-connectivity game}. This characteristic function was later generalised by \cite{Lindelauf:et:al:2013} to:
\begin{equation}
\label{equation:nu_fL}
\nu^{f}_G(S) =
\begin{cases}
f(S,G) & \textnormal{if } S \in \mathcal{C}(G) \\
0 & \textnormal{ otherwise, }  \\
\end{cases}
\end{equation}
\noindent where $f$ is an arbitrary function to the real numbers, allowing for any valuation of connected coalitions.

\cite{Amer:Gimenez:2004} combined the game $\nu^{\mA\mG}_G(S)$ with semivalues in order to rank individual nodes in terms of their importance to connectivity. \cite{Lindelauf:et:al:2013} considered the Shapley value of $\nu^{f}_G$, although---in principle---any other semivalue could also be used.

As a rather interesting application, \cite{Lindelauf:et:al:2013} used their centrality measure to analyse terrorist networks. However, \cite{Michalak:et:al:2013b} later argued that, for terrorist networks, this centrality measure overstates the importance of nodes for the connectivity of the network and thus is not suitable for this application.
In terms of complexity, \cite{Michalak:et:al:2013b} also showed that, even for the simplest variant of connectivity games, where $\forall_{S\in\mathcal{C}(G)} f(S,G) = 1$, computing the Shapley value of $\nu^{f}_G$ is \#P-Complete and therefore not tractable for larger networks.

In the next section based on connectivity, we discuss an alternative (and in some respects more sophisticated) notion of connectivity for game-theoretic payoff allocation in directed networks, which was introduced by \cite{Kim:Tackseung:2008}.

\subsection{Weak Connectivity due to \cite{Kim:Tackseung:2008}}\label{subsection:kim:tackseung}

\noindent \citefull{Kim:Tackseung:2008} introduce the notion of \emph{weak connectivity}, and use it to develop a payoff division scheme (that can be interpreted as a centrality measure) that is similar to the Myerson value. In fact, this division scheme is equal to the Myerson value if the network is undirected.
However, rather than requiring a path between any pair of nodes (i.e., strong connectivity), weak connectivity attempts to account for the fact that a single node in a digraph can ``gather'' the information from within a coalition. Such a node (which need not be unique) is referred to by the authors as the \textit{informational coordinator}. Let $E^*$ be the relation that represents the transitive closure of the relation defined by the edges of a digraph, $D(V, E)$. In other words, we have $E^*(s,t)$ if and only if there exists a path from $s$ to $t$. For $C \subseteq V$, if there exists a node $v \in C$ such that for all nodes $u \in C$ we have $E^*(u, v)$ (i.e., $v$ is an informational coordinator), then we say that $C$ is \emph{weakly connected}. By $C/E$ we denote the set of weakly connected components of $C \in V$ and define it as the coarsest partition of $C$ into weakly connected sets (i.e., such that no two weakly connected sets can be joined to form an even larger weakly connected set). Note that this partition need not be unique. If the partition is unique, then the authors define digraph restricted games in the following manner:

\[
\nu_D^{\mathcal{KT}}(S) = \sum_{C \in S/E} \nu(C).
\]

When the partition is not unique, the value of the digraph restricted game is defined as the maximum value over all partitions:

\[
\nu_D^{\mathcal{KT}}(S) = \max_{S/E \in \mathcal{KT}(S)} [\sum_{C \in S/E} \nu(C)],
\]

where $\mathcal{KT}(S)$ denotes the set of all possible partitions of the set $S$ into maximally weakly connected components. Next, the authors used the following allocation rule:

\[
\phi^{\mathcal{KT}}_i(\nu) = SV_i(\nu_D^{\mathcal{KT} }).
\]

When the network in question is undirected, then weak connectivity and strong connectivity are equivalent. Therefore, for undirected networks, this allocation is equal to the Myerson value. Its complexity for undirected networks is therefore also the same as that of the Myerson value. For directed networks, however, it is potentially worse (although this has not been studied in the literature to date).

	\begin{table}
		\centering
		\begin{tabular}{l|l}
			Not weakly connected & Weakly connected components \\ \hline
			$\{1,3\}$ & $\{1\}, \{3\}$ \\
			$\{1,4\}$ & $\{1\}, \{4\}$ \\
			$\{2,4\}$ & $\{2\}, \{4\}$ \\
			$\{1,2,4\}$ & $\{1,2\}, \{4\}$ \\
			$\{1,3,4\}$ & $\{1\}, \{3,4\}$ \\
			$\{1,5\}$ & $\{1\}, \{5\}$ \\
			$\{2,5\}$ & $\{2\}, \{5\}$\\
			$\{1,2,5\}$ & $\{1,2\}, \{5\}$ \\
			$\{1,3,5\}$ & $\{1\}, \{3,5\}$ \\
			$\{4,5\}$ & $\{4\}, \{5\}$ \\
			$\{1,4,5\}$ & $\{1\}, \{4\}, \{5\}$ \\
			$\{2,4,5\}$ & $\{2\}, \{4\}, \{5\}$ \\
			$\{1, 2,4,5\}$ & $\{1,2\}, \{4\}, \{5\}$ \\
			$\{1,3,4,5\}$ & $\{1\}, \{3,4,5\}$
		\end{tabular}
		\caption{The breakdown of the not weakly connected subgraphs of $\{1, 2, 3, 4, 5\}$ in Figure~\ref{figure:del:Pozo} into weakly connected components}
		\label{table:components}
	\end{table}

\begin{example}
	Consider again Figure~\ref{figure:del:Pozo} and the characteristic function $\nu(C) = |C|^2$. There are $31$ non-empty subsets of the set $\{1, 2, 3, 4, 5\}$. Of these, $17$ are weakly connected. As for the other $14$ subsets, they are listed in Table~\ref{table:components} along with their weakly connected components. In this case, all of the partitions are unique. None of these  subgraphs (except for singletons) would be considered connected in the classical sense, since we never have a situation where every node can communicate with every other node.

	The centralities of the nodes $v_1$, $v_2$, $v_3$, $v_4$, and $v_5$ according to \cite{Kim:Tackseung:2008} are $7.36667, 8.5, 9.33333, 7.46667$, and $7.46667$, respectively.
\end{example}

In the next section, we discuss an axiomatic approach to connectivity due to \cite{skibski2016attachment}. This will also be the last centrality based on connectivity that we discuss.

\subsection{Attachment Centrality due to \cite{skibski2016attachment}}

\cite{skibski2016attachment} developed an axiomatic approach to study connectivity in networks. In particular, they postulated the following axioms for a connectivity-based centrality measure:

\begin{itemize}
	\item \textbf{Locality:} The centrality of a node depends only on its connected component. In other words, let $C$ be the connected component of $v$ in some graph $G$. Then the centrality of $v$ in $G$ should be the same as its centrality in the subgraph of $G$ induced by $C$, $G[C]$
	\item \textbf{Normalisation:} The centrality of any node is inbetween the values $0$ and $|V|-1$, inclusive.
	\item \textbf{Fairness:} If we add an edge $(u,v)$ to $G$, then the magnitude of the resulting change to the centrality of $u$ is equal to the magnitude of the change to the centrality of $v$. This is the same axiom that is used to characterise the Myerson value.
	\item \textbf{Gain-loss:} Adding any edge to any connected graph, $G$, will not change the sum of the centralities of the nodes.
\end{itemize}

The authors show that these four axioms uniquely characterise the following centrality, which they call attachment centrality:
\[
A_v(G) = \sum_{S \in V \setminus \{v\} } 2 \beta(S,V) ( |\mathcal{K}(C)| - |\mathcal{K}(C \cup \set{v})| + 1 ),
\]
where $\beta(S,V) = \frac{|S|!(|V| - |S| - 1)!}{|V|!}$ and $\mathcal{K}(C)$ is the set of components of $G[C]$, the subgraph induced by $C$. The authors also note that $A_v(G)$ is equal to the Shapley value of the following game:
\[
\nu(C) = 2(|C| - |\mathcal{K}(C)|)
\]
In this, the approach of the authors to centrality based on connectivity is similar to Myerson, whereby it is defined by a set of axioms. This is highly desirable, since the axioms intuitively identify the centrality and therefore help identify those contexts where it is best to apply it.

With this, we conclude our review of centrality measures based on connectivity. In the next section, we discuss those that are not.

\section{Game-Theoretic Node Centrality Measures Not Based on Connectivity}\label{section:synergy}

\noindent The second general approach to game-theoretic network centrality places emphasis on choosing a characteristic function that evaluates the role of nodes according to their topological position and combining this with some solution concept for normal cooperative games (without networks).
In this approach, the connectivity of nodes does not need to have any direct bearing on their centrality, although this can still be expressed in the choice of the characteristic function. The general idea behind most of the centrality measures discussed in this section is to take some group centrality measure and treat it as a cooperative game on the set of nodes. Next, solution concepts such as the Shapley value, semivalues or others are used in conjunction with group centrality measures in order to yield a ranking of nodes.

This section is structured as follows. In Section~\ref{subsection:domination}, we discuss the work by  \cite{Brink:Borm:1994} and the follow-up work by \cite{Brink:Gilles:2000}, who defined a group centrality that reflects the \emph{dominance} of groups of nodes in directed graphs, and then proposed a Shapley value-based centrality that captures synergies within the network. This idea was rediscovered by \cite{Suri:Narahari:2008} for the application of information diffusion and influence propagation, and was extended by \cite{Michalak:et:al:2013} to a broader range of group centrality measures. Both publications are discussed in Section~\ref{section:michalak:jair}, where Shapley value degree and Shapley value closeness centrality measures are discussed. \cite{Michalak:et:al:2013} developed polynomial algorithms for computing these---and other---game-theoretic centrality measures. Continuing, in Section~\ref{section:betweenness} we discuss the work by \cite{Szczepanski:et:al:2012}, who proposed the Shapley value betweenness centrality and also developed polynomial-time algorithms for calculating it. 
In Section~\ref{section:owen:value:centrality}, we discuss a generalisation of the Shapley value approach to centrality that was developed by \cite{Szczepanski:et:al:2014}. This approach, which is based on the Owen value, applies to networks that have an extraneously defined community structure. Finally, in Section \ref{subsection:VL} we discuss the VL control measure, which shares some, but not all aspects of the other centrality measures discussed in this section.

\subsection{Centrality Based on Dominance due to \cite{Brink:Borm:1994}}\label{subsection:domination}

\noindent\cite{Brink:Borm:1994}  and \cite{Brink:Gilles:2000} propose a measure of \emph{dominance} in directed networks called the \emph{$\beta$-measure}.\footnote{\footnotesize A preliminary version of this article appeared as a working paper \cite{Brink:Borm:1994} and was later on extended by \cite{Brink:Borm:2002}.} Although this measure was apparently not intended to be game-theoretic in nature, the authors show that it coincides with the Shapley value of the \emph{score game} in networks (which is a generalisation of the so-called \emph{score measure}).
Moreover, the authors axiomatically characterize both the $\beta$-measure and score measure. Although it is not stated explicitly as such, the $\beta$-measure can be interpreted as a measure of centrality in networks. 
The remainder of this section formally introduces the score measure, the score game, and---finally---the $\beta$-measure.

In the context of this section, a directed network is interpreted as a representation of the dominance between nodes, whereby a directed edge $(v,u)$ represents the fact that $v$ dominates $u$. Let $E(v) = \{u \midd (v,u) \in D\}$ denote the set of nodes dominated by $s$, and let $E^{-1}(v) = \{u \midd (u,v) \in D\}$ denote the set of nodes that dominate $v$. The \emph{score measure}, then, is a function, $\sigma_D:V\to \mathbb{N}$ that assigns to every node, $v\in V$, the number of nodes that are dominated by $v$ (i.e., the outdegree of $v$). Formally:
\[
\sigma_D(v) = \left|E_D(v)\right|.
\]
This measure is well known in the literature and mentioned, for example, by \cite{Behzad:1979}. It is typically used to rank participants in a tournament, where a tournament is defined as an irreflexive digraph where for all $u,v \in V$ such that $u \neq v$ it holds that either $(u, v) \in D$ or $(v, u) \in D$. \cite{Brink:Gilles:2000} axiomatically characterized this measure, showing that it uniquely satisfies the following properties or axioms:

\begin{itemize}
	\item \textit{Score normalisation}: For every digraph, $D$, the sum of the scores of all nodes add up to $|E|$.
	\item \textit{Dummy-node}: If $v$ does not dominate any node, its score is 0;
	\item \textit{Symmetry}: For any $u, v \in V$ such that $E_D(u) = E_D(v)$ and $E_D^{-1}(u) = E_D^{-1}(v)$, the score of $u$ is equal to the score of $v$;
	\item \textit{Additivity over independent partitions}: The sum of the scores of nodes in $D$ is equal to the sum of the scores of nodes in the subgraphs generated by all the independent partitions of $D$ (where independent partition are such that for any node $u$, if the set of nodes that dominates $u$ is non-empty, then all of the nodes dominating $u$ belong to the same part in the partition), the score of any node $v$ is the sum of the values that $v$ receives in each part in the partition.
\end{itemize}

Inspired by the score measure, \cite{Brink:Borm:1994} introduced the \emph{score game} of a digraph $D$, which is a cooperative game with the set $V$ of players and the characteristic function---$\nu^{\mathcal{BB}}_D$---that generalises the score measure as follows:
\[
\nu^{\mathcal{BB}}_D(C) = \left|E^{\mathcal{BB}}_D(C)\right|,
\]
where $E^{\mathcal{BB}}_D(C) = \bigcup_{v \in C} E_D(v)$ (note that $E^{\mathcal{BB}}_D(C)$ may or may not contain members of $C$). In a sense, this is a version of group degree centrality for digraphs, where the out-degree of a coalition determines its score. Note, however, that for a coalition of nodes $C$, whereas group out-degree centrality does not usually count the out-neighbours of the members of $C$ that also belong to $C$ towards its value, the score game does.
%
%

	Finally, the $\beta$-measure can be defined as the Shapley value of the score game. Equivalently, the authors show that for every node $v$ in $D$, we have:
	

\[
\beta_v(D) = SV_v(\nu^{\mathcal{BB}}_D) =  \sum_{u \in E_D(v)} \frac{1}{|E_D^{-1}(u)|}.
\]

Intuitively, the measure indicates that a reward of $1$ is given for dominating any node. If a node, $v$, is dominated by multiple nodes, then this reward is equally distributed among all of the nodes that dominate it. \cite{Brink:Gilles:2000} axiomatically characterize their $\beta$-measure by the properties: dummy-node, symmetry, additivity over independent partitions, and dominance normalisation. Whereas we are familiar with the first four axioms, dominance normalisation is defined as follows:

\begin{itemize}
	\item \textit{Dominance normalisation}: For every digraph $D$, the sum of the scores of all players is equal to the number of players that are each dominated by at least one node (i.e., have an in-degree that is not equal to $0$).
\end{itemize}

{As a criticism of this axiomatisation, we note that it axiomatises a dominance relation through the use of axioms that inherently account for dominance. In this, the axiomatisation adds little new insight into understanding the measure.} To compute the $\beta$-measure of a node $v$, first the sets $E^{\mathcal{BB}}_D(v)$ and $|E_D^{-1}(u)|$ must be computed for any node $u$ dominated by $v$. This can be done in $O(|V| + |E|)$ time. \cite{Brink:Gilles:2000} later generalise the score measure and $\beta$-measure for weighted digraphs, however they do not show any link between these generalised measures and cooperative game-theory.

\subsection{The Top-k Nodes Problem and the Sphere of Influence}\label{section:michalak:jair}

	\noindent In this section, we review various versions of Shapley value degree and closeness centrality. The one thing that these centrality measures have in common is that they were proposed in order to model the sphere of influence of nodes. Following \cite{Brink:Borm:1994}, the second game-theoretic centrality based on degree was proposed by \cite{Suri:Narahari:2008}. The authors consider a cooperative game where the characteristic function is defined by the group degree centrality of each coalition. In other words, the value of a coalition of nodes is equal to the number of neighbours of this coalition. Next, the authors propose to use the Shapley value of this game as a centrality measure. \cite{Michalak:et:al:2013} follow by introducing a slightly different variant of a cooperative game based on degree centrality. In particular, the value of a coalition is equal to the size of its ``fringe'' set, which is defined as the set of its members and neighbours. Formally, the ``fringe'' of a subset $C\subseteq V$ is defined as follows:

\[\fringe(C) = \left\{v\in V : \big(v\in C\big) \textnormal{ or } \big(\exists u\in C :(u,v)\in E\big)\right\}.\] 
Building upon this, the authors define the following group centrality measure 
\begin{equation}\label{equation:g_1}
g_{1}(C) = |\textnormal{fringe}(C)|.
\end{equation} 
We see that this approach is very similar to the one introduced in the previous section. Just like \cite{Brink:Borm:1994} and \cite{Suri:Narahari:2008}, \cite{Michalak:et:al:2013} use the Shapley value of this game as a centrality measure. In terms of computation, whereas \cite{Suri:Narahari:2008} approximated their centrality measure using Monte Carlo techniques,  \cite{Michalak:et:al:2013} developed polynomial algorithms for the computation of the Shapley value of any node given the game $g_1$. \cite{Michalak:et:al:2013} proposed four additional centrality measures, which correspond to the Shapley values of nodes in the following games:



\begin{itemize}
	\item $\mathbf{g_2^{(k)}:}$ This game is inspired by the general-threshold model introduced by \cite{Kempe05influentialnodes}. It is parameterized by a threshold value---$k$---and the general idea is that the value of a coalition is equal to the number of players that are either in $C$ or have at least $k$ neighbors in $C$. In a sense, whereas $g_1$ assumes that controlling any neighbour of a node, $v$, is sufficient to control $v$, $g_2^{(k)}$ assumes that at least $k$ neighbours of $v$ must be controlled in order to influence it. Formally, $g_2^{(k)}(G)(C)=|\{v\in V: (v\in C) \textnormal{ or } (|E(v)\cap C|\geq k)\}|$.
	\item $\mathbf{g_3:}$ Unlike the previous games, $g_3$ is defined over weighted graphs.
	Under this game, the value of a coalition, $C$, is equal to the number of nodes that are within a certain ``cutoff distance'' from $C$. Formally, $g_3(G)(C)=|\{v\in V : \mathit{dist}(v,C)\leq d_{\mathit{cutoff}})\}|$. 
	Intuitively, the cutoff distance can be interpreted as the ``radius'' within which any node can influence other nodes. We note that $g_1$ is a special case of $g_3$, where $d_{\mathit{cutoff}} = 1$ and the weight of each edge is $1$.
		\item $\mathbf{g_4:}$ According to this game, the value of a coalition $C$ is the sum over all nodes not belonging to $C$ of a positive-valued, non-increasing function, $f: \mathbb{R} \rightarrow \mathbb{R}$ of the distances from $C$ to those nodes.TOMASZTODO In other words, this is exactly the generalised form of group closeness, $\psi^{CL}_{f}(G)(S)$, discussed in Section \ref{ch:3-preliminaries}.
	We note that $g_3$ is a special case of $g_4$ where $f(\dist(v,C))=1$ if $\dist(v,C)\geq d_{\mathit{cutoff}}$, and $f(\dist(v,C))=0$ otherwise.
	\item $\mathbf{g_5:}$ This game is also defined over weighted graphs, but unlike $g_3$ and $g_4$, the weight of an edge is not interpreted as its length, but rather its \emph{power of influence}.  The influence of a coalition, $C$, on a node, $v$, is defined as the sum of all influences induced upon $v$ by the members of $C$. Now, for every node $v$, let us introduce a threshold, $W_{\mathit{cutoff}}(v)$, that is necessary to influence it. The value of a coalition, $C$, is then equal to the number of nodes that are either in $C$ or are influenced by $C$. Formally, $\nu_5(C)=|\{v\in V: (v\in C) \textnormal{ or } (\sum_{u\in C}\mathit{\omega(u,v)}\geq W_{\mathit{cutoff}}(v))\}|$, where $\omega(u,v)$ is the weight of the edge between $u$ and $v$.
	Note that $g_2^{(k)}$ is a special case of $g_5$ where $W_{\mathit{cutoff}}(v)=k$ for all $v\in V$ and $\omega(u,v)=1$ for all $(u,v)\in E$.
\end{itemize}

\cite{Michalak:et:al:2013} proposed exact and efficient algorithms for computing the Shapley value-based centrality measures based on $g_1$ to $g_4$. As for the Shapley value of $g_5$, the authors propose an approximation algorithm. 
A summary of the computational complexity of their algorithms  for computing the Shapley value of the five games is presented in Table~\ref{front_table}.
Additionally, the authors evaluate the effectiveness of their algorithms on two real-life networks: an infrastructure network that represents the topology of the Western States Power Grid and a collaboration network from the field of astrophysics.

\setlength{\tabcolsep}{0.2em}
\begin{center}
	\begin{table}[t]
		\begin{center}
			\begin{tabular}{l l l}
				\hline
				\hline
				Game  & Complexity & Accuracy \tabularnewline
				\hline
				$g_{1}$ & $O(|V|+|E|)$ & exact \tabularnewline
				\hline
				$g_{2}^{(k)}$  &  $O(|V|+|E|)$  &exact \tabularnewline
				\hline
				$g_{3}$  &  $O(|V||E|+|V|^{2}\textnormal{log}|V|)$  &exact \tabularnewline
				\hline
				$g_{4}$ & $O(|V||E|+|V|^{2}\textnormal{log}|V|)$ &exact \tabularnewline
				\hline
				$g_{5}$ & $O(|V||E|)$ &approx. \tabularnewline
			\end{tabular}
		\end{center}
		\caption{Games considered by \cite{Michalak:et:al:2013} and their computational results.}
		\label{front_table}
	\end{table}
\end{center}

\cite{Suri:2010} considered another characteristic function for the application of information diffusion. In order to express the fact that information has reached a node, we will say that a node has been \textit{activated}. Once a node has been activated, it can share this information with its neighbors. These types of situations can appear in ad campaigns, social networks or politics. Within this context, \cite{Suri:2010} are concerned with determining which nodes are the most important in disseminating information.

To model dissemination of information, \cite{Suri:2010} base their characteristic function on the number of activated nodes in the Linear Threshold model. Next, the Shapley value is used as a measure of centrality. The Linear Threshold model consists of a weighted graph with active and/or inactive nodes, each of which has a threshold value. Nodes are activated in iterative fashion by their neighbors provided that the sum of the weights of an inactive node's neighbors surpasses its threshold value. The value of a coalition according to the characteristic function is then defined to be the number of active nodes in a graph after information has been disseminated in so many rounds, such that another iteration would yield no more active nodes. \cite{Suri:2010} use Monte Carlo approximations in order to compute this centrality.

	\cite{Matejczyk:et:al:2014} study the appropriateness of the Shapley value to the influence maximisation problem. They propose two new game-theoretic algorithms for this problem: the discount Shapley value algorithm and a refinement of the Local Directed Acyclic Graph (LDAG) algorithm. They conclude that the greedy LDAG algorithm is currently the best solution, however their Shapley value LDAG algorithm works almost as well and can be implemented in a parallel map-reduce fashion. Moreover, the Shapley value LDAG algorithm produces a larger cover (i.e., the set of nodes that is influenced) than any other previously studied game-theoretic solution.

We have now surveyed various models of domination and information diffusion and game-theoretic centrality measures that account for it. In doing so, we introduced three variations of Shapley value degree centrality \cite{Brink:Borm:1994,Suri:Narahari:2008,Michalak:et:al:2013} and a generalised Shapley value closeness centrality (based on $g_4$). The natural next step is to apply the Shapley value to group betweenness centrality, which presents a useful measure of network flow. We address this in the next section.

\subsection{The Shapley Value Betweenness Centrality due to \cite{Szczepanski:et:al:2012}}\label{section:betweenness}
In the previous section we introduced game-theoretic network centrality measures that were developed by applying the Shapley value to variations of both group degree centrality and group closeness centrality. In this section, we introduce the Shapley value betweenness centrality, which was proposed by \citeauthor{Szczepanski:et:al:2012} \cite{Szczepanski:et:al:2012,Szczepanski:2016}. 
%
%

	Formally, the centrality is defined as the Shapley value for the cooperative game defined by group betweenness centrality, $\nu^B$, which we defined in Section~\ref{ch:3-preliminaries}. \cite{Szczepanski:et:al:2012} show the centrality can be reformulated as follows:
	
	\begin{equation}
	SV_v(\nu_B) = \sum_{s \neq v} \sum_{t \neq v} \bigg( \frac{\sigma_{s,t}(v)}{\sigma_{s,t}\dist(s,t)} + \frac{2 - \dist(s,v)}{2\dist(s,v)}\bigg),
	\end{equation}
	
	\noindent where $\sigma_{s,t}$ is the number of shortest paths in the graph $G$ between the nodes $s$ and $t$, and $\sigma_{s,t}(v)$ is the number of those paths that contain $v$. The authors show that this formulation naturally leads to a polynomial algorithm for the computation of their centrality measures.

\begin{figure}[thbp]
	\center
	\hspace*{-0.7cm}\includegraphics[width=15cm]{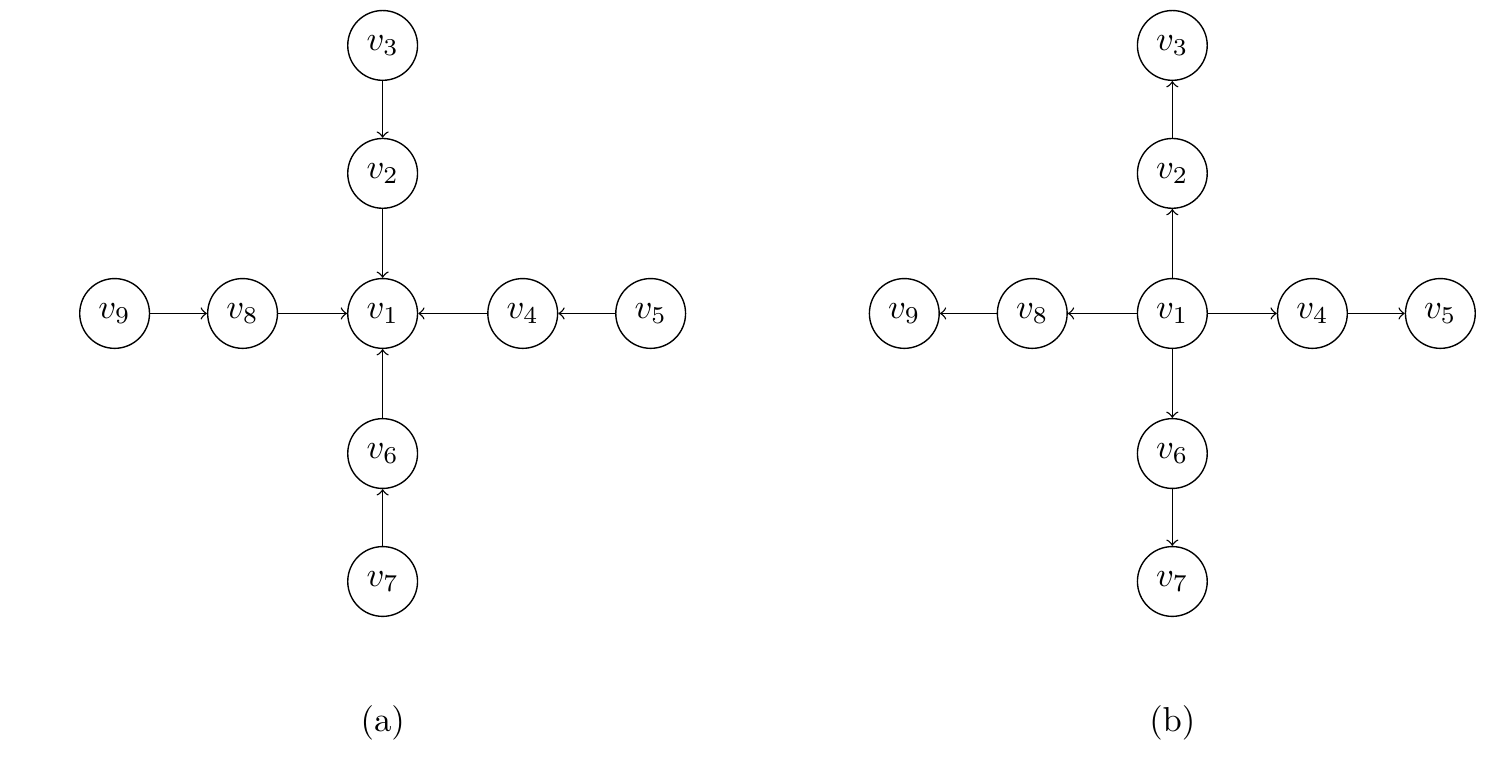}
	\caption{Two sample networks showing one of the difficulties with node centrality in digraphs. Depending on the application, any nodes could be considered as ``central.''
	}
	\label{figure:2star}
\end{figure}

	\begin{example}
		Consider Figure~\ref{figure:2star} (a) and (b). Some of the centrality measures we have seen thus far would place most importance on either the outer nodes ($v_3, v_5, v_7$ and $v_9$) or the inner node ($v_1$). Both the standard betweenness centrality and its Shapley value extension, however, place most emphasis on the intermediary nodes---$v_2, v_4, v_6, v_8$. In fact, the central and outer nodes in the networks all have centrality $0$ according to standard and Shapley value betweenness centrality, since none of them are intermediary nodes in any paths. Consider also the Banzhaf path-based centrality due \cite{Grofman:Owen:1982}, which considers a coalition ``winning'' when it forms a simple path, and focuses on players that have ``swing'' votes (i.e., whose removal makes it impossible for the two ends of the path to communicate).  
		Since all the possible paths in Figure~\ref{figure:2star} (a) and (b) are shortest paths, the Banzhaf path-based  centrality \cite{Grofman:Owen:1982} will result in the exact same ranking as betweenness centrality (assuming that the beginning and end nodes of a path are not allowed to have swing votes).
	\end{example}

In the next section, we discuss a generalisation of the centrality measures from this section that considers a more complex setting, where nodes belong to communities.

\subsection{The Owen Value-based Centrality due to \cite{Szczepanski:et:al:2014}}\label{section:owen:value:centrality}

\noindent In this section we discuss the work of \citefull{Szczepanski:et:al:2014}, who developed the first measure of centrality that takes into account the \emph{community structure} of the underlying network. The measure is based on a generalisation of the Shapley value known as \emph{the Owen value}---a well-known solution concept from cooperative game theory that focuses on games with \textit{a priori}-given unions (i.e., cooperative arrangements) of players. Moreover, the authors develop a class of solution concepts that they call coalitional semivalues, which are a generalisation of the Owen value and semivalues. We refer the reader to Section~\ref{ch:3-preliminaries}, where we introduced these concepts.

\cite{Szczepanski:et:al:2014} propose to use coalitional semivalues in combination with group degree centrality as a measure of centrality in graphs with a community structure. Much like in the general Shapley-value approach used by \cite{Suri:Narahari:2008} and \cite{Michalak:et:al:2013}, coalitional semivalues can be used to modify classical centrality measures to account for synergies. The authors provide a polynomial algorithm to compute their measure.

The main advantage of this approach is that it takes into account the additional information afforded by the community structure in determining the centrality of nodes. The authors argue that the community to which a node belongs should also impact its centrality. That is, if a node's community is weak, then this should impact the centrality of the node negatively, and if it is strong, then it should impact it positively. In particular, the authors use their centrality measure to analyse a coauthorship network, where communities are identified by the various venues (journals, conferences, etc.) where the nodes (i.e., authors) have published. The main advantage of the Owen value approach is that it accounts for the fact that being a top contributor to weak venues does not necessarily make one a good researcher, whereas having fewer publications in top venues may be more advantageous. We discuss this further in Section \ref{section:applications}.

Building upon the Owen value, \cite{Tarkowski:2016} introduce a new class of solution concepts for cooperative games with overlapping community structures and combine them with general group closeness centrality to develop a robust and general centrality measure for graphs where the communities of nodes can overlap.

The next section introduces the last measure from the literature---the VL Control Measure, as introduced by \cite{Hendrickx:et:al:2009}. Although this measure also focuses on synergies, it does so in a much different manner than the other measures in this section.

\subsection{VL Control Measure for symmetric networks due to \cite{Hendrickx:et:al:2009}}\label{subsection:VL}

\noindent In this section, we discuss the VL control measure (where VL stands for Vorobev Liapounov) due to \cite{Hendrickx:et:al:2009}. This measure---much like those surveyed in Section \ref{section:michalak:jair}---also has applications in influence propagation and information diffusion. 
To illustrate the measure, let us analyse a fictitious context considered by the authors. Imagine that some amount of a noxious substance is hidden in the nodes of a network. The object is to eliminate all of the substance.
To do so, a \textit{searcher} assigns $x_v$ resources to each node $v$, with the constraint that $\sum_{v \in V} x_v = 1$. That is, the searcher has $1$ unit of resource that can be freely divided up among the nodes. The probability of finding the substance at any given node $v$ is $y_v = x_v + \sum_{(v, u) \in E} x_u$. That is, resources can search for the noxious substance in the node they are assigned to and in its neighbors. The probability of removing all of the substance, then, is $\prod_{v \in V} y_v$. Alternatively, we can think of allocating resources in order to spread information in a network, or any similar setting. Finally, the VL control measure of a given node, $v$, is the amount of resources that needs to be placed at $v$ according to the strategy that maximises the probability of removing all of the substance. In other words, the centrality of nodes is the solution to the maximisation problem $\max(\prod_{v \in v} y_v)$ with the constraints $y_v = x_v + \sum_{(v, u) \in E} x_u$ for all $v$, $\sum_{v \in V} x_v = 1$ and $\forall_{v \in v} x_v \geq 0$. By placing resources to appropriate nodes we want to maximise the probability that all of the substance will be removed, while staying within the restrictions of the problem (e.g., not using more resources than we have at our disposal). Note that this solution need not be unique.

\begin{example}
	Consider the nodes in Figure~\ref{figure:Belau}. First, let us think how we can raise the probability of removing all of the substance up from $0$. Putting any amount of resources at any one node will not suffice to achieve this. Clearly, then, resources must be placed in at least $2$ nodes. Moreover, placing any resources at $v_1$ or $v_5$ makes no sense, since a better effect can be achieved by placing the resources in nodes $v_2$ or $v_3$, respectively (since they additionally cover $v_4$). The answer we need is the solution to the maximisation problem $r_2 * (r_2 + r_3 + r_4) * (r_2 + r_3 + r_4) * (r_2 + r_3 + r_4) * r_3$ subject to $\sum_{i = 1}^5 r_i = 1$, where $r_i$ is the amount of resources placed in node $v_i$ (we have already informally established that $r_1 = 0$ and $r_5 = 0$). Again, we can correctly surmise that no resources should be placed in $v_4$, since placing them in $v_2$ or $v_3$ would achieve the same effect while also protecting $v_1$ and $v_5$. The answer to the problem, then, is placing half of the resource in $v_2$, and the other half in $v_3$, yielding a centrality ranking vector of $0, 0.5, 0.5, 0, 0$.
	
\end{example}

At first glance, it does not look like this measure has anything to do with cooperative game theory. However, the centrality of each node according to this measure turns out to be a \textit{proper Shapley value} of the game $\nu(S) = |\{v \in S : E(v) \subseteq S\}|$, i.e., the number of nodes in a coalition that do not have connections to nodes outside of it. The Harsanyi dividends for this game are easily expressed as follows: {$\Delta_{\nu}(S) = |\{v \in S : E(v) = S\}|$}. To continue, we will introduce the \textit{weighted Shapley value}, which will lead us to the definition of the proper Shapley value.

\textbf{The Weighted and Proper Shapley Values:} Harsanyi Dividends can be used to develop a payoff division scheme, where the value is not shared equally among players. One such division scheme is called the weighted Shapley value. In this payoff division scheme, it is possible to give more importance to certain players (i.e., to disregard the Symmetry axiom). The idea is that a weight is assigned to every player through a weight vector $\omega = \{\omega_1, \omega_2, \ldots, \omega_i, \ldots, \omega_n \}$, where $\omega_i$ is the weight of player $i$.\footnote{\footnotesize  Note that the weights do not need to sum up to $1$, since the weighted Shapley value considers the relative power of players in every coalition in accordance with the weights. For example, for the coalition $C = \{2, 3, 5\}$, player $2$ would be assigned $\frac{\omega_2}{\omega_2 + \omega_3 + \omega_5}$ power, player $3$ would be assigned $\frac{\omega_3}{\omega_2 + \omega_3 + \omega_5}$ power and player $5$ would be assigned $\frac{\omega_5}{\omega_2 + \omega_3 + \omega_5}$ power.} The weighted Shapley value with weight vector $\omega$ is then:

	\begin{equation}
	\label{WEIGHT_SV}
	SV^{\omega}_i(\nu) = \sum_{C \in \set{C \midd C \subseteq I \text{ and } i \in C} }  \frac{\omega_i}{\sum_{j \in C} \omega_j} \Delta_{\nu}(C).
	\end{equation}

Whenever for all players $i \in I$ the case is that $SV^{\omega}_i(\nu) = \omega_i$, then $SV^{\omega}_i(\nu)$ is referred to as a \textit{Proper Shapley value} for $\nu$.

\textbf{Equivalence of the concepts:} The equivalence of the maximisation problem to the proper Shapley value was showed by \cite{vorobev:liapounov:1998}. In particular, the authors showed that the solution to the following maximisation problem is a proper Shapley value of $\nu$:
\[
\max_{x \in \mathcal{S}^I} \prod_{S \subseteq I} (\sum_{j \in S} x_j)^{\Delta_{\nu}(S)},
\]
where $\mathcal{S}^I$ is the unit simplex and $x_j$ is the $j$'th element of the vector $x$. This can be rewritten as:
\begin{align*}
&\max \prod_{S \subseteq I} (\sum_{j \in S} x_j)^{\Delta_{\nu}(S)}, \nonumber\\
&\sum_{i = 1}^n x_i \leq 1,\nonumber \\
&\forall_{i \in I} x_i \geq 0. \nonumber
\end{align*}

For the game $\nu$ this can be applied as follows:
\[
\max_{x \in \mathcal{S}^N} \prod_{S \subseteq N} (\sum_{j \in S} x_j)^{|\{i \in S : E(i) \subseteq S\}|},
\]
which is equivalent to the maximisation problem:
\[
\max_{x \in \mathcal{S}^I} \prod_{i \in I} \sum_{j \in E(i)} x_j,
\]
which is the VL control measure that we introduced at the beginning of this section.

%
%
%
%

Unfortunately, \cite{NONLINEAR:PROGRAMMING} showed that there is no polynomial algorithm for non-linear programming even for a quadratic polynomial unless $\text{P}=\text{NP}$. Moreover, there is no polynomial time approximation for this problem (even with a very poor guarantee) unless $\text{P}=\text{NP}$. {However, even though the VL control measure can be reduced to an instance of a nonlinear programming problem, the actual computational complexity of the VL control measure has yet to be extensively studied.}


\begin{figure}[thbp]
	\center
	\includegraphics[width=15.5cm]{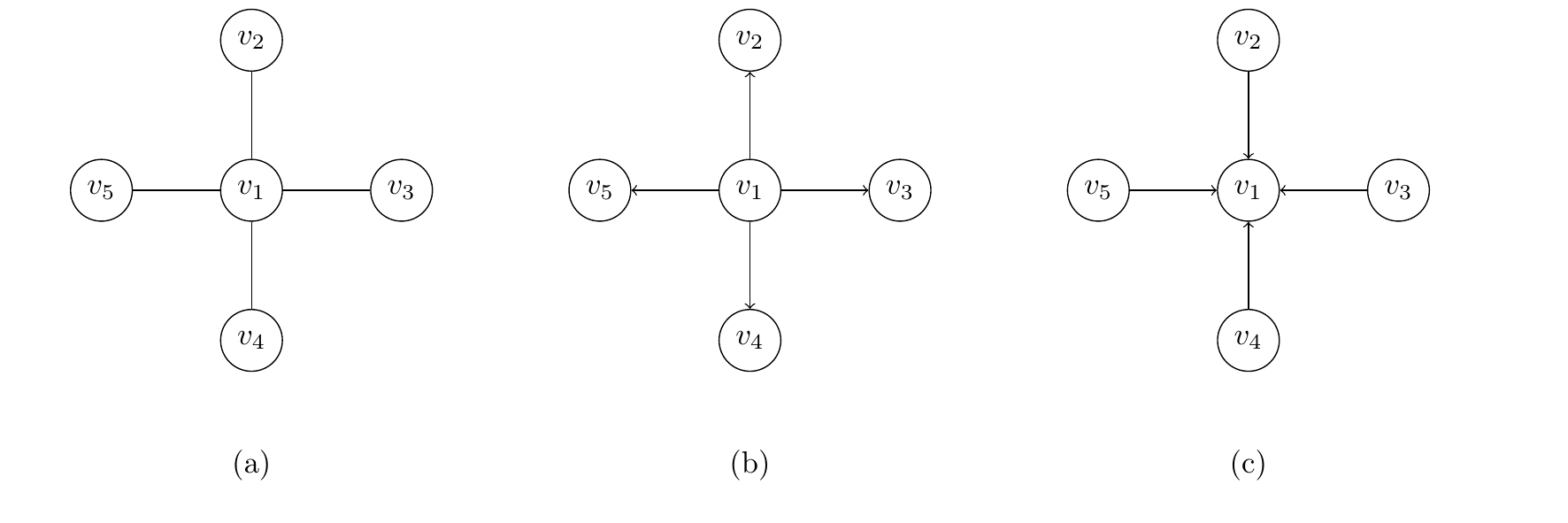}
	\caption{Three sample networks showing one of the difficulties with node centrality in digraphs. Most centrality measures agree that $v_1$ is the most central node in network (a). However, if a centrality measure admits that node $v_1$ is central in network (b), then it should not be central in (c) (and vice versa).
	}
	\label{figure:first:last:example}
\end{figure}

\begin{example}\label{example:3star}
	Consider now Figure~\ref{figure:first:last:example}. We reach a fundamental question for centrality in directed networks: in which of these networks (all of which are ``stars'') is the center node---$v_1$---central? The answer is usually straightforward in the undirected graph (a). Both \cite{del:Pozo:2011} and \cite{Hendrickx:et:al:2009} (with a strictly convex game), for example, agree that $v_1$ should be the most central node. However, if we assume that $v_1$ is the most central in digraph (b) (as it is according to \cite{Hendrickx:et:al:2009}), then we require central nodes to be able to travel, or send information to other nodes in the network. For this same reason, $v_1$ cannot be the most central node in digraph (c). On the other hand, if we assume $v_1$ to be the most central node in digraph (c) (as it would be according to  \cite{del:Pozo:2011}), then our application requires central nodes to be reachable from other nodes in the network. Clearly, then, $v_1$ cannot be the most central node in digraph (b), since no other node can reach it.
\end{example}

\subsection{A More General Model}

Whereas the focus of most work has been the computation of a single solution concept for a single group centrality, \cite{Szczepanski:tarkowski:2015} present a generic framework for defining group centralities. If the group centrality is defined within this framework, any semivalue can be computed in polynomial time using a generic algorithm proposed by the authors. Interestingly, the framework also facilitates the complexity analysis and the development of algorithms for classes of group centrality measures (e.g., parametrised measures) at once, rather than analysing every measure individually.


\section{Applications}\label{section:applications}

\noindent One weakness of the literature on game-theoretic network centrality is that it has not been applied very extensively. This is perhaps due to a lack of algorithms proposed and the computational difficulties inherent in the field. We list below a few of the applications for which game-theoretic network centrality has been used.

\cite{Suri:Narahari:2008} apply their Shapley value-based degree centrality to a co-authorship network of $8 361$ high-energy Physics 
researchers. They show that their centrality for the top-$k$ nodes problem achieved better results than a well-known algorithm from the literature---the maximum degree heuristic.

\cite{Skibski:et:al:2014} and \cite{Michalak:et:al:2015} use the Myerson value in order to identify key nodes in terrorist networks. The authors argue that this helps understand the hierarchy of such organisations and can facilitate efficient deployment of investigation resources. 

	\cite{Szczepanski:et:al:2014} applied their Owen value degree centrality to a citation network. The unique feature of this approach is that it is able to take into account the importance of the community to which a node belongs in evaluating its centrality. Unfortunately, however, the method cannot be applied when nodes belong to more than one community (i.e., communities overlap). The citation network that was studied consists of $2$~$084$~$055$ publications and $2$~$244$~$018$ citation relationships. A total of $22$~$954$
	unique communities representing journals, conference proceedings
	or single book titles were identified by the authors by using basic text mining techniques. The authors found that the Owen value-based degree centrality has a significant advantage over other centralities in ranking the authors. This is shown through a comparison to two other centrality measures---weighted degree centrality and Shapley value-based degree centrality. The authors show that some nodes are significantly less powerful according to the Owen value centrality, since the communities to which they belong are weak. They argued that the Owen value-based centrality is able to account for the fact that being a strong author in only weak journals does not make one a strong author in general. 

\cite{Szczepanski:2016} proposed to use semivalue betweenness centrality in order to protect networks. The idea is that random node failures may close lines of communication in a network, and semivalues can help rank nodes in terms of how important it is to protect them. They find that their methods perform favourably when compared to a number of other centrality measures.

\cite{Matejczyk:et:al:2014} use the Shapley value and Banzhaf index on games with networks to solve the top-$k$ nodes problem. They find that their method performs comparatively (although a little worse) to the state-of-the-art.

\citeauthor{Keinan:2004} \cite{Keinan:2004} use the Shapley value in order to rank how important sections of the brain are for certain cognitive functions. Given the complex interactions between sections of the brain, they argue that the Shapley value can account for the synergies achieved by them.

\citeauthor{Skibski:et:al:2016:KCG} \cite{Skibski:et:al:2016:KCG} proposed the first game-theoretic centrality measure that is based on the extension of the Shapley value to games with externalities \cite{Skibski:et:al:2017} and  advocated its usefulness to the analysis of the well-known board game of Diplomacy.

\citeauthor{Michalak:et:al:2015:social:capital} \cite{Michalak:et:al:2015:social:capital} argue that game-theoretic centralities can be considered as novel measures of social capital that address two key deficiencies of standard measures. Firstly, while the standard  measures focus separately on various types of social capital (e.g., on either individual or group social capital),  game-theoretic centralities (e.g., those based on the Owen value) can be used to measure interactions between such different types of social capital. Secondly, network-based standard social capital measures focus solely on the network topology; hence, they do not take into account various additional information about the nodes, groups and connections between them.  In contrast, such additional information can be embodied in the characteristic function upon which game-theoretic centralities are built upon.

Game-theoretic centralities become power indices when applied to weighted voting games restricted by a graph. For instance, the Myerson value is the extension of the Shapley-Shubik index to graph-restricted weighted voting games.  
It is argued that restricted weighted voting games better model reality as we cannot expect that cooperation of any subset of political parties is always feasible \cite{Fernandez:et:al:2002}. For instance, it is hard to expect that a new entrant to the German Bundestag---the AfD party---will be treated by CDU/CSU in the same way as SPD. Unfortunately, power indices are computationally challenging \cite{Chalkiadakis:et:al:2011,matsui_voting}. Hence, various authors have recently proposed dedicated algorithms for computing power indices in special cases of graph-restricted weighted games \cite{Fernandez:et:al:2002,Skibski:et:al:2015:pseudo,Benati:et:al:2015,Skibski:Yokoo:2017}.

\citeauthor{Narayanam:et:al:2014} \cite{Narayanam:et:al:2014} introduce a game-theoretic centrality measure based on the Shapley value in which the characteristic function is especially designed to promote ``gatekeepers''---nodes that play an important role in connecting their communities with the remainder of the network. They analyse two application of their centrality measure: community detection and limiting the spread of misinformation over the network.

\cite{Szczepanski:2015b} use a class of solution concepts for pairs of players in cooperative games that is based on semivalues---semivalue interaction indices (introduced in Section \ref{ch:3-preliminaries}). They pair this class of solution concepts with group $k$-degree centrality (where the value of a coalition, $C$, is the number of nodes at least $k$ nodes away from it) in order to analyse the similarity of nodes in networks. This has two main applications: link prediction and community detection. In a missing information scenario, where random edges are removed from a network, the authors show that their approach is competitive when compared with the state-of-the-art methods in the literature. \cite{Szczepanski:2016b} continued this research and develop a measure of node similarity for players that belong to different communities. They base their measure on coalitional semivalues and refer to it as coalitional semivalue interaction indices, which we also introduced in Section \ref{ch:3-preliminaries}. The authors also pair this solution concept with group $k$-degree centrality and find that it is well-suited to predicting whether edges exist between nodes that belong to different communities (i.e., the inter-links prediction problem). This is more complicated than the standard link-prediction problem, since there are usually very few edges from nodes in one community to those in any other community, meaning there is less information to base the similarity measure on. 


\section{Conclusions}\label{section:conclusion}
\noindent We have organised the literature on game-theoretic network centrality measures into two categories and presented the relevant centrality measures and included a short discussion of the computational complexity of each measure. We note that whereas the measures based on connectivity are generally intractable, those that are present in the literature that are not {based on connectivity}, are often computable in polynomial time. This is a surprising result, given the complexity of the solution concepts that these measures have adopted from cooperative game theory. We also mention the work due to \cite{Skibski:2017}, who organised game-theoretic centrality measures according to their axiomatic properties.

\section*{Acknowledgements}
Tomasz
Michalak and Michael Wooldridge were supported by the European Research Council
under Advanced Grant 291528 (“RACE”). Talal Rahwan, and Tomasz Michalak were supported by the Polish National
Science Centre grant DEC-2013/09/D/ST6/03920.

\clearpage
\section*{Appendix A: Summary of Main Notation}

\begin{longtable}{ll}
$V$ & {The set of nodes.} \\
$E$ & {The set of edges.}\\
$\omega: E \rightarrow \mathbb{R}$ & {Weight function that assigns to each edge a real number.}\\
$G=(V,E)$ & {Undirected graph with node set $V$ and edge set $E$.} \\
$D=(V,E)$ & Directed graph. \\
$G=(V,E,\omega)$ & {Undirected weighted graph with weight function $\omega$.} \\
$D=(V,E,\omega)$ & {Directed weighted graph.} \\
$C$ or $S$ & {A coalition.} \\
$I$ & {A set of players of a cooperative game.} \\
$n$ & {The number of players in a cooperative game.} \\
$\nu: 2^I \rightarrow \mathbb{R}$ & {A characteristic function.} \\
$\nu(C)$ & {A value of the coalition, where $\nu$ is characteristic function.}\\
$\Delta_\nu(C)$ & The Harsanyi Dividend of $C$ in the game $\nu$.\\
$\beta(k)$ & {The probability that a coalition of size $k$ is chosen.} \\
$SV_{i}(\nu)$ & {The Shapley value of the player $i$ in game $\nu$.} \\
$u,v,s,t \in V$ & {The node from the set $V$.} \\
$deg(v)$ & {Degree of the node $v$.} \\
$E(v)$ & {Set of neighbours of node $v \in V$.} \\
$E_{IN}(v)$ & {Set of in neighbours of node $v \in V$.} \\
$E_{OUT}(v)$ & {Set of out neighbours of node $v \in V$.} \\
$\dist(v,u)$ & {The distance from the node $v$ and $u$.} \\
$MC(u,v)$ & {Marginal contribution that node $u$ makes through node $v$.} \\
$\psi$ & {A group centrality measure.} \\
$\phi$ & {A payoff division scheme.} \\
$\CS$ & {A community structure (cover) of the set $V$.}\\
$Q$ & {A community such that $Q \in  CS$.}\\
$T$ & {A coallition of communities, i.e. $T \subseteq  CS$.}\\
$S_{\nu}(C,i,j)$ & {The synergy between $i$ and $j$ in the context of coalition $C$.}\\
\end{longtable}
\newpage

\bibliographystyle{plainnat}
\bibliography{referencesconsolidated}

\end{document}